\documentclass{article}

    \PassOptionsToPackage{numbers, compress, sort}{natbib}

 \usepackage[main, final]{neurips_2026}

\usepackage[utf8]{inputenc} 
\usepackage[T1]{fontenc}    
\usepackage{hyperref}       
\usepackage{url}            
\usepackage{booktabs}       
\usepackage{amsfonts}       
\usepackage{nicefrac}       
\usepackage{microtype}      
\usepackage{xcolor}         

\hypersetup{
  colorlinks  = true,      
  urlcolor    = RoyalBlue, 
  linkcolor   = RoyalBlue, 
  citecolor   = RoyalBlue   
}
\usepackage[svgnames,dvipsnames,table]{xcolor}
\usepackage{url}
\usepackage{booktabs}  

\usepackage{amsmath,bm}
\usepackage{graphicx}
\usepackage{enumitem}
\usepackage{subcaption}
\usepackage{soul} 
\usepackage{multirow}
\usepackage{wrapfig}
\usepackage{pifont}
\usepackage{fix-cm}

\title{TACO: Towards Task-Consistent Open-Vocabulary Adaptation in Video Recognition}

\author{%
  Minghao Zhu \qquad Xiao Lin \qquad Mengxian Hu \qquad Xun Zhou \\
  \textbf{Liuyi Wang} \qquad \textbf{Xiaoyan Qi} \qquad \textbf{Chengju Liu}\thanks{Corresponding author.} \qquad \textbf{Qijun Chen} \\
  Tongji University, Shanghai, China\\
  \texttt{\{zmhh\_h, linx\_xx, humengxian, zhouxun,wly} \\ 
  \texttt{2331853, liuchengju, qjchen\}@tongji.edu.cn}
}

\begin{document}

\maketitle

\vskip -0.4cm
\begin{abstract}
\vskip -0.1cm
Adapting CLIP for open-vocabulary video recognition necessitates a delicate balance between newly acquired video knowledge and the pretrained generalization.
While existing studies pursue this generalization-specialization trade-off with additional regularizations or constraints, we argue that they overlook the deviation of representations beyond the fine-tuning data distribution, resulting in suboptimal adaptation effects.
We believe such deviation is inherited from the inconsistency between the fine-tuning and evaluation objectives, where model optimization is restricted to the known training distribution but evaluated on unseen ones.
In this paper, we introduce \emph{TACO}, a simple yet effective framework to mitigate the potential negative effects induced by this inconsistency.
Our key insight is that adaptation should preserve OOD-relevant alignment beyond the training distribution. 
To this end, we propose \emph{Relative Structure Distillation}, which regularizes the relative geometry of the representation space and suppresses harmful alignment shift during training. 
We further decouple the representation space from the optimization space with a lightweight specialization projection, allowing task-specific adaptation without directly overspecializing the representations used at test time.
\emph{TACO} establishes state-of-the-art performance on diverse benchmarks under cross-dataset and base-to-novel settings.
Code will be released at \url{https://github.com/ZMHH-H/TACO}.
\end{abstract}
\vspace{-0.2cm}

\section{Introduction}
\label{Introduction}
Open-vocabulary learning aims to identify novel visual concepts specified by language vocabularies unseen during training phase.
This setting has been propelled by the rapid progress of large-scale Vision–Language Models (e.g., CLIP~\cite{CLIP}, ALIGN~\cite{ALIGN}, Florence~\cite{Florence}), which learn aligned multimodal representations from massive image-text pretraining.
With their strong generalization to unseen visual concepts, these foundation models are highly practical in real-world applications and have been widely adapted to various downstream tasks~\cite{CoOp, ViLD, Catseg, CLIPose,FIMA}.
Inspired by this progress, recent studies have explored adapting CLIP for general video recognition~\cite{XCLIP, ViFiCLIP, MoTE}, avoiding the substantial cost of collecting large-scale video-text data and pretraining from scratch.

While CLIP provides a strong starting point for visual understanding, effective adaptation for open-vocabulary video recognition remains challenging - specifically in empowering models to accurately capture the temporal dynamics encoded in videos of unknown categories.
The key challenge is to introduce video-specific knowledge while preserving the pretrained generalization, a problem commonly framed as the generalization/specialization trade-off.
To strike this balance, recent studies (e.g., Open-VCLIP~\cite{OpenVCLIP}, FROSTER~\cite{FROSTER}) typically seek a middle ground between the two aspects by incorporating carefully designed regularizations during fine-tuning.
Despite their promising results, we argue that these methods overlook representation variation beyond the training distribution, as their optimization objectives are strictly confined to the fine-tuning data distribution (i.e., ID space).

\begin{figure}[t]
\begin{center}
\centerline{
    \includegraphics[width=0.7\linewidth]{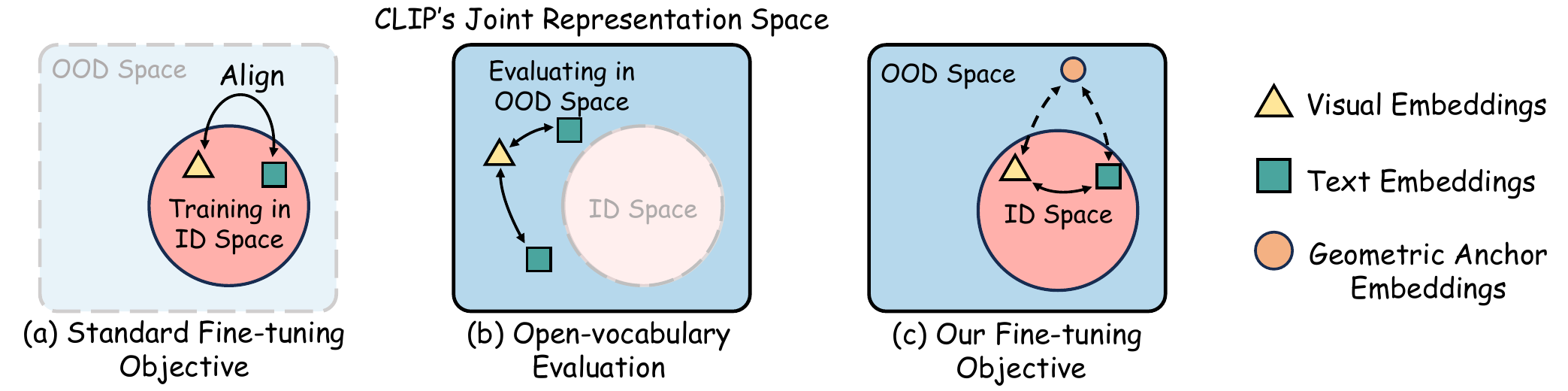}}
\vskip -0.19cm
\caption{(a) Standard fine-tuning aims to align visual and text embeddings within the known training distribution (ID Space), but neglects the deviation in the out-of-distribution space.
While (b) the open-vocabulary model is evaluated in the OOD space, (c) our fine-tuning objective can effectively maintain the relative structure across the entire embedding space and improve the generalization.
}
\label{Fig::Teaser}
\end{center}
\vskip -0.7cm
\end{figure}

Intuitively, without explicit constraints on the out-of-distribution (i.e. OOD) space, visual and text representations may drift in different directions, thereby disrupting their cross-modal alignment.
Under the open-vocabulary setting, such alignment disruption can directly undermine the model's generalization to videos from unseen categories.
In practice, since the fine-tuning dataset is limited in both scale and diversity, this inconsistency between fine-tuning and evaluation objectives is prevalent in existing adaptation paradigms, as illustrated in Figure~\ref{Fig::Teaser}.
We revisit this inconsistency to better understand what must be preserved for open-vocabulary generalization.
Our analysis suggests that OOD generalization still largely relies on CLIP’s original semantic space, whereas standard fine-tuning often induces representation drift and cross-modal alignment shift beyond the training distribution, leading to degraded generalization.


Building upon these findings, we present \emph{TACO}, a simple yet effective framework for task-consistent open-vocabulary video adaptation, and formulate a more explicit adaptation principle.
Our key insight is that the adaptation process should properly regularize the entire representation space, thereby aligning more closely with the evaluation objective and curbing harmful alignment shift in OOD space.
To this end, we propose \emph{Relative Structure Distillation}, which ensures the consistent structure of the entire representation space throughout fine-tuning in a data-independent manner.
Unlike existing distillation objectives that focus on matching within training distribution, our objective connects the ID space with a constructed OOD space, thereby effectively suppressing drift in OOD regions.
We further show that simple random geometric anchors sampled on the CLIP hypersphere are sufficient to provide effective global structural references.
Second, we propose a lightweight \emph{Specialization Projection} that decouples the representation space from the optimization space, so that task-specific adaptation can occur in a disposable branch rather than directly overspecializing the representations used at test time.
These two designs directly follow our analysis: RSD protects the global alignment structure beyond the training categories, while the specialization projection reduces the tendency of the cross-entropy objective to overfit the shared representation space. 
Overall, \emph{TACO} is concise, scalable, and easily integrated into different adaptation pipelines, delivering significant improvements.
Extensive experiments show that TACO consistently improves open-vocabulary generalization and establishes new state-of-the-art performance across multiple datasets and backbone scales.

Our main contributions can be summarized as follows:
\begin{itemize}[noitemsep,topsep=0pt,leftmargin=*]
\item We introduce \emph{TACO}, a simple yet effective framework for open-vocabulary video adaptation. 
\emph{TACO} aims to mitigate the potential negative effects induced by the representation deviation beyond the training distribution, a factor not directly addressed in prior work.
\item We revisit the objective inconsistency in existing fine-tuning paradigms, analyze its impact on preserving pretrained generalization, and formulate a more explicit adaptation principle (\S~\ref{Sec::Analysis}).
\item We present Relative Structure Distillation, which effectively preserves generalization by keeping the consistent structure of OOD space without explicit OOD supervision.
We further propose a generic representation-space decoupling scheme to mitigate overspecialization (\S~\ref{Sec::Methodology}).
\item 
\emph{TACO} sets state-of-the-art performance on cross-dataset and base-to-novel settings across multiple benchmarks.
Comprehensive ablations further demonstrate the effectiveness of our method (\S~\ref{Sec::Experiments}).

\end{itemize}



\section{Analysis: Revisiting Open-Vocabulary Adaptation for Video}
\label{Sec::Analysis}
\vskip -0.02cm
\subsection{Preliminary: Adapting CLIP for Video Recognition}
To transfer the powerful generalization of the CLIP model~\citep{CLIP} to the video domain, recent studies have fine-tuned its joint embedding space using video-text data and obtained promising results~\citep{XCLIP,ViFiCLIP,MoTE}.
We briefly describe the standard fine-tuning paradigm in this context.
Consider a CLIP-based video learner composed of a visual encoder $f_{\theta_v}(\cdot)$ and a text encoder $f_{\theta_t}(\cdot)$, where $\theta_v$ and $\theta_t$ are the parameters of each encoder.
Given a video $V \in \mathbb{R}^{T \times H \times W \times 3}$ with $T$ frames and the corresponding text description $C$ embedded in a set of pre-defined templates (e.g., \emph{"A video of []"}), we extract the visual embedding $\bm{v} \in \mathbb{R}^{D}$, text embedding $\bm{c} \in \mathbb{R}^{D}$, and compute their similarity $\operatorname{sim}(\bm{v}, \bm{c})$ as:
\begin{equation} 
\label{Eq::eq1}
\bm{v} = f_{\theta_v}(V), \enspace \bm{c} = f_{\theta_t}(C), \enspace \operatorname{sim}(\bm{v}, \bm{c}) = \frac{\langle\bm{v}, \bm{c}\rangle}{\left\|\bm{v}\right\| \left\|\bm{c}\right\|},
\end{equation}
where $D$ is the dimension of the joint embedding space.
During fine-tuning, video-specific knowledge is injected by encouraging the video embedding to align with the matched category text embedding while separating it from other categories.
Formally, the training objective can be written as:
\begin{equation} 
\label{Eq::eq2}
\mathcal{L}_{CE} = \mathbb{E}_{\scriptscriptstyle (V,C)\sim\mathcal{D}}[\, CE(\,\operatorname{sim}(\bm{v}, \bm{W}),\, onehot(C)\,)\,],
\end{equation}
where $\mathcal{D}$ is the fine-tuning dataset, $\bm{W} \in \mathbb{R}^{N \times D}$ denotes the text embeddings of $N$ training categories, $CE(\cdot, \cdot)$ is the cross-entropy loss with softmax operation, and $onehot()$ denotes the one-hot encoding.

\subsection{Inconsistency Between Open-vocabulary Adaptation and Evaluation}
\label{Sec::Analysis::Inconsistency}

Following the above fine-tuning paradigm, great efforts have been made to adapt CLIP for open-vocabulary video recognition~\citep{OpenVCLIP,FROSTER,MoTE,OpenMeDe}.
These methods typically conduct fine-tuning on the  Kinetics-400~\citep{K400} dataset, and the derived model is expected to generalize well on test data with unseen categories $C_{test} \in \mathcal{S}_{test}$, where $|\mathcal{S}_{ft} \cap \mathcal{S}_{test}| < |\mathcal{S}_{ft} \cup \mathcal{S}_{test}| $.
$\mathcal{S}_{ft}$ and $\mathcal{S}_{test}$ denote the fine-tuning and test vocabularies.
However, the standard fine-tuning objective is optimized restrictively within the in-distribution (ID) training space, neglecting the basic setting of evaluating open-vocabulary tasks in the OOD space.
In this section, we analyze the potential negative impact of such inconsistency in preserving generalization, supported by empirical evidence.


\paragraph{A closer look at preserving generalization.}
\begin{wrapfigure}[10]{r}{0.45\textwidth}
    \vspace{-3.5ex}
    \begin{center}
    \centerline{
        \includegraphics[width=1.0\linewidth]{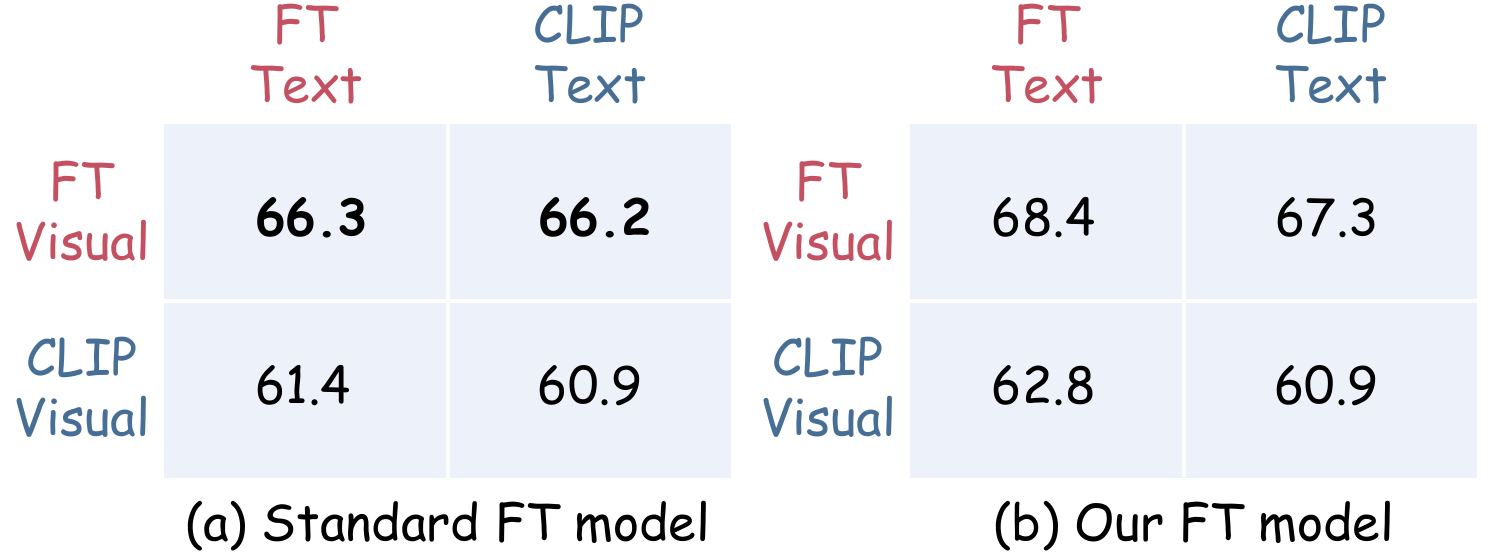}}
        \vspace{-1.0ex}
    \caption{Replacing the encoders of the standard fine-tuning model and our model with the original CLIP encoders. 
    }
    \label{Fig::Exchange}
    \end{center}
    \vskip -1.0cm
\end{wrapfigure}
To better understand the essence of preserved generalization, we investigate the impact of each fine-tuned encoder on generalization by replacing it with the corresponding CLIP's original encoders during evaluation.
We show the harmonic mean of zero-shot performances in Figure~\ref{Fig::Exchange} (a).
Interestingly, replacing the fine-tuned text encoder with CLIP's text encoder has almost no impact on the results (66.3\% vs. 66.2\%), which suggests that the OOD generalization of the adapted model is grounded in CLIP's original semantic space.
However, fine-tuning inevitably leads to deviations in the OOD representation space (evidenced in Figure~\ref{Fig::Relativesim}), as the representations of the test data are not orthogonal to the ID subspace spanned by the training data~\citep{LP-FT}.
This deviation may further distort the representation alignment between visual and text modalities in OOD space and result in diminished generalization capability.
Based on these findings, we hypothesize that \emph{maintaining consistent representation alignments in OOD space is crucial for preserving generalization.}




\paragraph{Characterizing representation deviation in OOD Space.}
\begin{wrapfigure}[14]{r}{0.45\textwidth}
    \vspace{-3.5ex}
    \begin{center}
    \centerline{
        \includegraphics[width=1.0\linewidth]{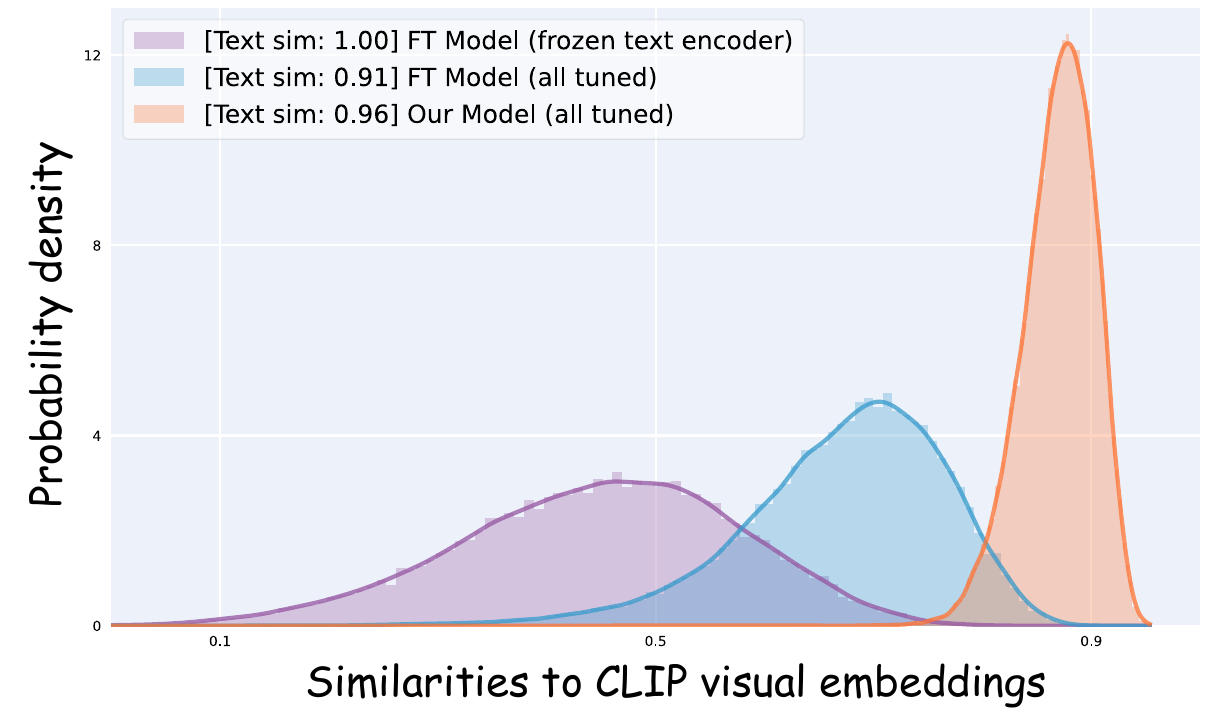}}
        \vspace{-1.5ex}
    \caption{Similarity distributions of visual embeddings between the CLIP model and various fine-tuned models in OOD space (UCF, HMDB, and K600).
    }
    \label{Fig::Relativesim}
    \end{center}
    \vskip -0.8cm
\end{wrapfigure}
To support the above hypothesis, we investigate how representations deviate in OOD space during fine-tuning.
Specifically, Figure~\ref{Fig::Relativesim} visualizes the similarity distributions between visual embeddings from the CLIP model and various fine-tuned models, with the averaged overall text embedding similarity shown in the legend.
From the figure, a pronounced deviation can be observed in visual embeddings for standard fine-tuning paradigms.
Interestingly, enabling the text encoder to be updated can effectively suppress this deviation.
This phenomenon reveals that the text encoder serves as a critical regularizer to \emph{relieve the overfitting in visual representations, rather than learning new textual knowledge}.
However, there is still a noticeable deviation for both fine-tuned visual and text embeddings that degrades the OOD generalization.
We believe such deviation should be further curbed during fine-tuning, which was overlooked in previous studies.


\paragraph{Quantifying alignment shift in adaptation.}
\begin{wrapfigure}[13]{r}{0.45\textwidth}
    \vspace{-3.2ex}
    \begin{center}
    \centerline{
        \includegraphics[width=1.0\linewidth]{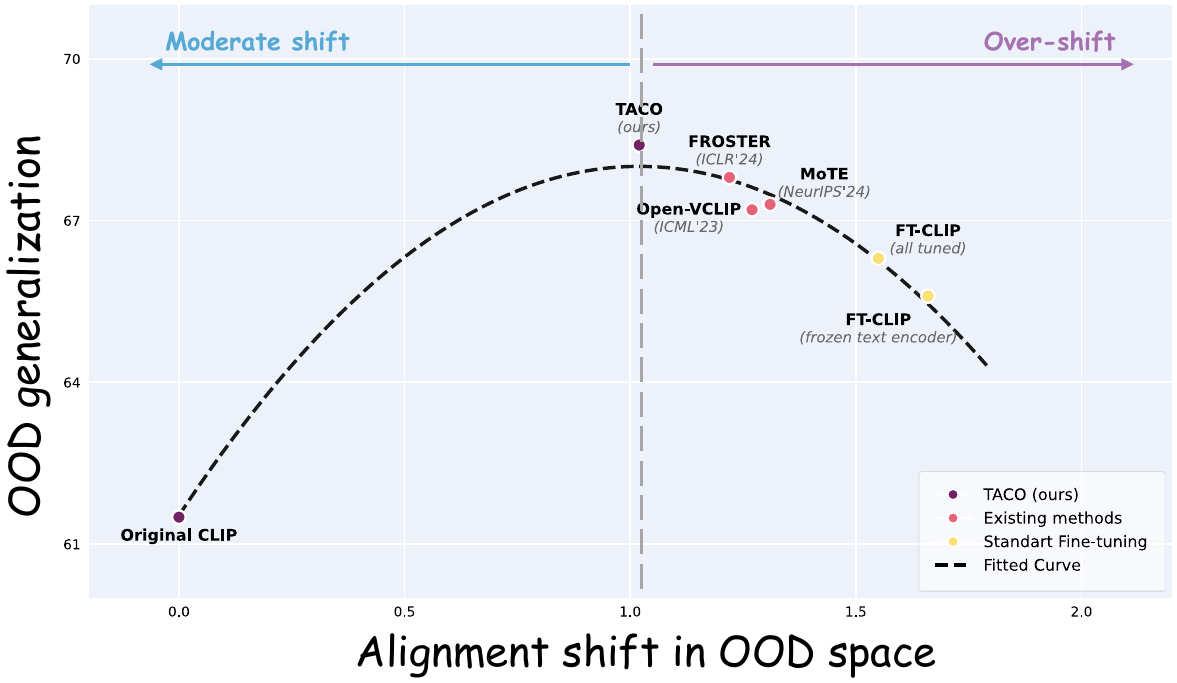}}
        \vspace{-1.5ex}
    \caption{Alignment shift $D_{\text{KL}}$ and generalization performance of various adapted models in OOD space (UCF, HMDB, and K600)
    }
    \label{Fig::AlignmentShift}
    \end{center}
    \vskip -0.8cm
\end{wrapfigure}
Beyond the deviation of individual representations in OOD space, a more critical issue is whether such deviation further disrupts the original cross-modal alignment.
We therefore quantify the resulting alignment shift and examine its relationship with the generalization performance HM$_{ood}$.
Specifically, we use the visual-text similarity matrix of the original CLIP in OOD space as the reference alignment, and measure the shift of an adapted model by the KL divergence $D_{\text{KL}}$ between their similarity matrices. 
A larger $D_{\text{KL}}$ indicates a greater departure from the original alignment.
As shown in Figure 4, a moderate degree of alignment update benefits generalization by bridging the image–video domain gap, whereas \emph{excessive shifts correlate with degraded generalization due to cross-modal disruption}.
This finding offers a new lens for interpreting the efficacy of existing methods by highlighting their role in protecting OOD alignment.
However, existing methods generally suffer from over-shifting, as they overlook the importance of constraining the variation in OOD space.


\section{Methodology}
\label{Sec::Methodology}

\subsection{Motivation and Intuition}
\begin{figure*}[t]
\begin{center}
\centerline{
    \includegraphics[width=0.8\linewidth]{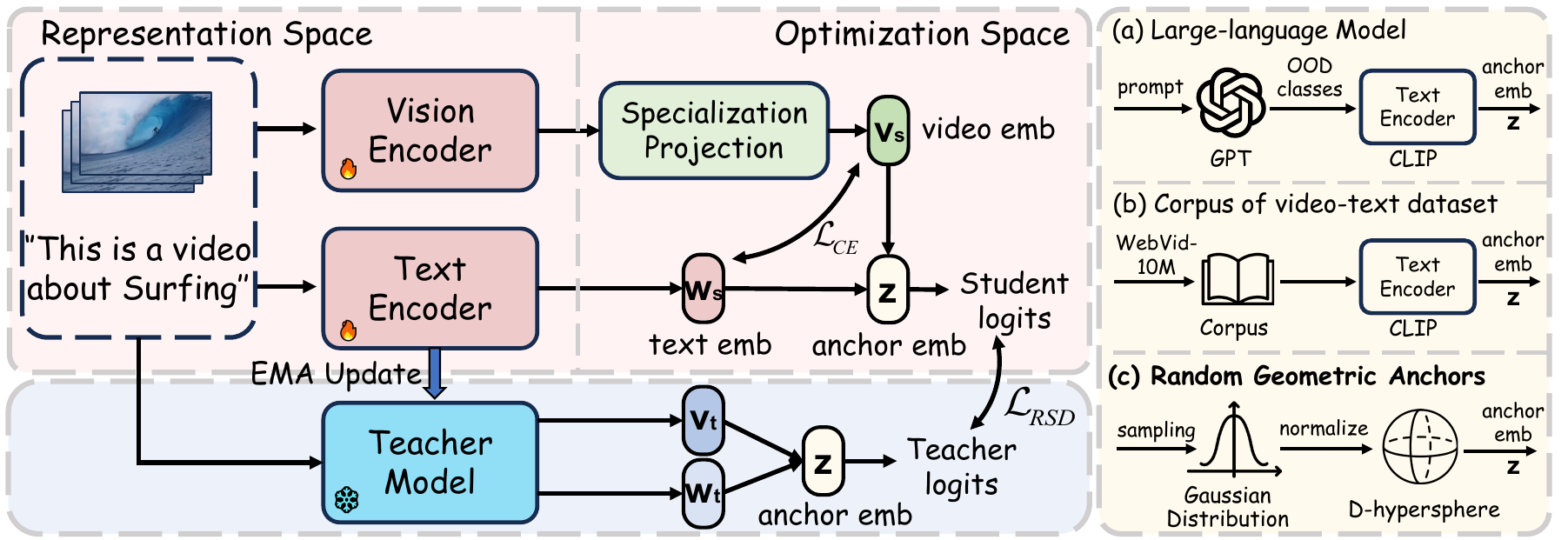}}
\vspace{-4pt}
\caption{An overview of the TACO framework (left) and the source of geometric anchors (right).}

\label{Fig::Method}
\end{center}
\vskip -1.0cm
\end{figure*}

Building upon the above analysis, we argue that maintaining consistent representation alignments in the OOD space is critical for generalizable VLM adaptation. 
Although existing methods can shape a well-structured ID space, their optimization objectives remain confined to the training distribution, offering limited constraint over representation variation beyond it and thus leading to suboptimal adaptation. 
A straightforward remedy would be to explicitly model the OOD space using real paired OOD vision-language data.
In practice, however, this is extremely challenging, as it requires collecting large amounts of real OOD data and incurs substantial computational overhead. 
This challenge inspires us to seek an indirect yet effective alternative: rather than explicitly modeling OOD semantics, we regularize the alignment of the entire representation space by preserving the \textbf{relative structure} between the ID and constructed OOD spaces. 
In this way, we can bypass the expensive collection and computational costs required by direct OOD modeling. 
This intuition naturally leads to our \emph{Relative Structure Distillation}, a novel regularizer that enforces geometric structural consistency with the pre-trained teacher beyond the training distribution, mitigating harmful alignment shifts without explicit OOD supervision, as presented in Figure~\ref{Fig::Method}.
From a geometric perspective, RSD can be viewed as random-projection-based relational regularization: the anchors probe structural changes without requiring advance knowledge of which directions correspond to unseen semantic concepts.

\subsection{Relative Structure Distillation}
In Section~\ref{Sec::Analysis::Inconsistency}, we reveal that the preserved generalization of an adapted model is rooted in CLIP’s original semantic space. 
Building on this insight, the core challenge during fine-tuning is to maintain consistency with the pretrained reference alignment in the OOD space.
Naturally, this challenge can be cast as a knowledge distillation problem. 
However, existing distillation strategies—ranging from standard Kullback-Leibler (KL) divergence~\cite{VL2V} and L2 matching~\cite{FROSTER} to relation-based variants~\cite{CRCD}—are fundamentally limited in this setting. 
Their distillation objectives typically focus on matching within the training distribution since the distillation data is identical to the training data, leaving them ineffective in suppressing deviations beyond it.
In contrast, our proposed \emph{Relative Structure Distillation} introduces principled solutions to the above challenge via refinements in loss formulation, distillation data, and optimization space.


\paragraph{Formulation of Relative Structure Distillation.}
Our design inspiration is to extend the regularization scope to the entire representation space by establishing connections between the training distribution and \emph{a constructed OOD representation space}.
Formally, given a video and a set of categories, we obtain the normalized teacher visual embedding $\bm{v_t} \in \mathbb{R}^{1\times D}$ and teacher text embeddings $ \bm{W_t} \in \mathbb{R}^{N \times D}$, along with the student embeddings $\bm{v_s}, \bm{W_s}$ of the same dimensions.
The teacher model shares the same structure as the student model.
Then we sample $M$ normalized anchor representations $\bm{Z} \in \mathbb{R}^{M \times D}$ from the constructed OOD space.
We take $\bm{Z}$ as geometric anchors to enforce the consistency of its relative relationships with the visual and text embeddings in ID space:
\begin{equation}
\label{Eq::eq5}
\definecolor{backcolor}{RGB}{240, 245, 254}
\mathcal{L}_{RSD}
= D_{KL}\Big(
\sigma\big(
\fcolorbox{black}{backcolor}{$\bm{v}_t \bm{Z_{}}^\top$}\,
\fcolorbox{black}{backcolor}{$\bm{Z_{}}\bm{W}_t^\top$}
/\tau
\big)
\,\big\|\,
\sigma\big(
\fcolorbox{black}{backcolor}{$\bm{v}_s \bm{Z_{}}^\top$}\,
\fcolorbox{black}{backcolor}{$\bm{Z_{}}\bm{W}_s^\top$}
/\tau
\big)
\Big),
\end{equation}
where $\sigma$ denotes the softmax operation, $\tau$ is a temperature parameter.
Here, our goal is to protect the \emph{geometric structure} of the entire representation space rather than modeling the truly out-of-distribution semantics.
The geometric anchors $\bm{Z}$ capture the relations with the visual embedding and text embedding through the terms $\bm{v} \bm{Z_{}}^\top$ and $\bm{Z_{}}\bm{W}^\top$. 
During fine-tuning, if the specialized CE loss overemphasizes the agreement between $\bm{v}$ and $\bm{W}$ in the ID space, their relations to the constructed OOD space may drift away. The anchor-mediated relations $\bm{v} \bm{Z_{}}^\top\bm{Z_{}}\bm{W}^\top$ can capture such relative structural changes and steer the student back toward a teacher-consistent geometry.


Appendix~\ref{Appendix::Sec::Theory} provides further theoretical support for the random-anchor formulation of RSD. Standard KL matching can leave cross-modal relation discrepancies invisible to ID logits. For any nonzero class-pair relation discrepancy in this null space, isotropically sampled anchors produce a positive expected squared response, providing additional geometric measurements beyond direct logit matching.

\paragraph{Construction of the OOD space.}
The geometric anchors $\bm{Z}$ are designed to serve as geometric reference points in the representation space, providing a geometric scaffold for preserving the structure of the entire space.
In this regard, the anchor embeddings sampled from the constructed OOD space do not need to be semantically associated with the training samples.
However, for such anchors to serve as meaningful geometric references, the constructed OOD space should
(\textbf{\romannumeral1}) reside in the same joint representation space as the model (\emph{compatibility}) and 
(\textbf{\romannumeral2}) ideally span the entire semantic space (\emph{coverage}).
One possible way is to sample from a large authentic corpus. 
For example, leveraging the textual descriptions in existing large-scale video-text datasets (e.g., WebVid-10M~\citep{WebVid}) or prompting large-language models~\citep{GPT-4o} to generate potential video categories.
But the sampled texts may not uniformly span the semantic space and require additional computations for embedding.

To address this, we propose \emph{Random Geometric Anchors}, which directly sample anchor embeddings from the unit $D$-hypersphere of the CLIP model, where $D$ denotes the embedding dimension. 
Specifically, we draw samples from the standard Gaussian distribution $\mathcal{N}(0, I)$ and project them onto the unit hypersphere by normalization. 
Owing to the rotational invariance of the Gaussian distribution, the resulting anchors are uniformly distributed on the hypersphere. 
As previous work has shown that CLIP’s representation space is confined within an extremely narrow conical region on the $D$-hypersphere~\cite{MindTheGap}, the hypersphere can be regarded as a generalized OOD space, and the probability of overlap between the sampled anchors and the ID space is practically negligible during fine-tuning. 
Therefore, although these anchors do not carry authentic semantic meaning, they naturally satisfy the desired properties of compatibility and coverage, and can effectively characterize the relative structure between ID and OOD spaces.
Moreover, these anchors should not be viewed as mere noise, but rather as geometric reference points that support structure preservation. 
This is analogous to preserving a map's layout with reference landmarks: even if the landmarks themselves are not meaningful destinations, they still help maintain the relative structure of the entire space.

The utility of these anchors does not require high cosine similarity to individual semantic directions. For a uniformly sampled unit anchor $z$, $\mathbb{E}[zz^\top]=I/D$, giving no preferred direction in expectation. Resampling introduces new geometric probes across iterations, with their sensitivity to relation discrepancies analyzed in Appendix~\ref{Appendix::Sec::Theory}.

\paragraph{Design of the teacher model.}
While the frozen CLIP provides a highly stable alignment reference as the teacher model, we observe that its distillation supervision may hinder the model’s ability in learning new knowledge.
In our work, the teacher model is designed as an exponential moving average (EMA) model with its parameters slowly updated at each iteration as $\tilde{\theta}  = \alpha\tilde{\theta} + (1-\alpha)\theta$, where $\alpha$ is a momentum coefficient, $\tilde{\theta}$ and $\theta$ are the parameters of the teacher and student models.
This mechanism ensures the teacher model remains consistent while evolving with new knowledge. 
\\The overall learning objective can be defined as follows, with $\lambda=0.4$ by default:
\begin{equation} 
\label{Eq::eq6}
\mathcal{L} = \mathcal{L}_{CE}+\lambda\,\mathcal{L}_{RSD},
\end{equation}

\subsection{Decoupling of the Representation Space}
In standard fine-tuning, the representation space is typically identical to the optimization space, as the output visual embeddings are directly used to compute the training objective.
However, the narrow supervision from a limited set of text categories tends to pull CLIP’s originally open and transferable visual representations toward a skewed subspace specialized for the training distribution. 
As a result, the very space that should generalize to unseen categories becomes directly exposed to task-specific bias. 
While Relative Structure Distillation explicitly regularizes the geometry of the representation space, it does not prevent the video-specific CE loss from directly specializing that space. 
Therefore, we further introduce a mechanism to protect the representation space from such direct specialization pressure, thereby mitigating harmful representation drift.

To this end, we propose to decouple the representation space from the optimization space by simply inserting a lightweight \emph{Specialization Projection} between them. 
Specifically, given a video input $V$, we first obtain its visual representation 
$f_{\theta_v}(V)$ and then transform it into the optimization space with a projection network $h(\cdot)$ as
$
\label{Eq::eq3}
\bm{\hat{v}} = h(f_{\theta_v}(V)) + f_{\theta_v}(V).
$
Here $\bm{\hat{v}} \in \mathbb{R}^{D}$ is the projected visual embedding used for optimization, while $f_{\theta_v}(V)$ is retained as the underlying representation.
To delicately balance the effective decoupling effect while keeping the alignment between text and pre-projected visual embeddings, we adopt a residual linear projection with zero-initialized parameters~\cite{LoRA} to avoid abrupt shifts at the start of the fine-tuning. 
We further remove non-linear activations, as they tend to cause harmful alignment changes when the text encoder is jointly optimized, and use a larger learning rate for the projection head than for the visual encoder to strengthen the decoupling effect.

During evaluation, we discard the projection head for more generalizable representations.
Our goal here is not merely generic overfitting reduction, but to prevent the representation space from being directly specialized by the CE objective. 
Unlike conventional regularizations such as weight decay or dropout, our decoupling strategy is an architectural bias that allows specialization to occur in a disposable optimization branch, rather than in the representation space that must generalize to unseen categories.
Thus, it is particularly well-matched to our goal of protecting the space structure.

\section{Experiments}
\label{Sec::Experiments}


\vskip -0.1cm
\subsection{Experimental Setup}
\label{Sec::Experiments::Setup}


\paragraph{Evaluation Protocols.}
We thoroughly evaluate our method with two common protocols: \emph{Cross-dataset} and \emph{Base-to-novel} evaluation~\cite{FROSTER, ViFiCLIP}.
\emph{(\textbf{\romannumeral1}) Cross-dataset:}
In this setup, the model is trained on Kinetics-400~\citep{K400} and then evaluated on other datasets with out-of-distribution vocabulary, including UCF-101~\citep{UCF101}, HMDB-51~\citep{HMDB51}, and Kinetics-600~\citep{K600}.
For UCF and HMDB, we evaluate on both the full dataset and three validation splits.
For K600, we adopt the three splits provided by \citep{ER}. 
Each split contains 160 categories sampled from 220 unseen categories.
\emph{(\textbf{\romannumeral2}) Base-to-novel:}
In this protocol, a dataset is divided into two disjoint category sets: base classes and novel classes.
The model is tuned on base classes and evaluated on both base and novel classes.
Evaluation datasets including K400, UCF, HMDB, and Something-Something v2~\citep{SSv2}.
During inference, we sample 3 temporal clips with a center crop (i.e. $3\times1$ views) per video, with each view consisting of 8 frames.

\paragraph{Implementation Details.}
We use the CLIP~\citep{CLIP} pre-trained ViT-B/16 and ViT-L/14 models in our experiments.
During fine-tuning, we sparsely sample 8 frames as the video input.
The pre-processing includes random cropping and resizing to the size of $224 \times 224$, along with random horizontal flips and grayscale.
We adopt AdamW~\citep{adamw} optimizer with a weight decay of 0.2. 
The initial learning rate is set to $3.75 \times 10^{-6}$ with a total batch size of 192, following a half-period learning rate decay.
For specialization projection parameters, the initial learning rate is increased to $5 \times 10^{-5}$.
Furthermore, we set the anchor embedding number $M$ to 200 and the momentum coefficient $\alpha$ to 0.9998. 
The geometric anchors are resampled at each training iteration.
\emph{Please see supplementary for more details.}

\begin{table*}[t]
\captionsetup[subtable]{skip=2.5pt}
\normalsize
\renewcommand\arraystretch{0.97}
\setlength{\tabcolsep}{2.0mm}
\definecolor{backcolor}{RGB}{230, 239, 250}
\caption{
Ablations on the components and key details. We report the cross-dataset performance on UCF, HMDB, and K600 split1, using the ViT-B/16 network. UCF$_f$ and K600$_f$ denote the results of freezing the text encoder in fine-tuning. Default settings are colored in \sethlcolor{backcolor}\hl{blue}.}

\vspace{-11pt}
\label{Tab::Ablation}
\begin{center}
\fontsize{8.6pt}{10pt}\selectfont
\begin{minipage}{0.32\textwidth}
\setlength{\tabcolsep}{0.3mm}
\begin{subtable}{1.0\linewidth}
\centering
    \caption{Effects of each components.} 
    \label{Tab::AS::components}
    \begin{tabular}{cc|ccc}
    Spec. Proj.     & $\mathcal{L}_{\mathrm{RSD}}$   &  UCF      & HMDB       & K600    \\
    \specialrule{0.7pt}{0pt}{0pt}
                    &                       & 82.8       & 52.0       & 73.4   \\    
    \ding{51}       &                       & 83.7       & 53.2       & 74.8   \\ 
                    & \ding{51}             & 84.5       & 52.9       & 75.2   \\
    \cellcolor{backcolor}\ding{51}       & \cellcolor{backcolor}\ding{51}             & \cellcolor{backcolor}\textbf{84.9}       & \cellcolor{backcolor}\textbf{54.2}       & \cellcolor{backcolor}\textbf{75.7}   \\
    \end{tabular}%
\end{subtable}%
\end{minipage}
\hfill
\begin{minipage}{0.32\textwidth}
\setlength{\tabcolsep}{0.8mm}
\begin{subtable}{1.0\linewidth}
\centering
    \caption{Source of geometric anchors.}
    \label{Tab::AS::generation}
    \begin{tabular}{c|cccc}
    Methods       & UCF            & HMDB          & K600         \\
    \specialrule{0.7pt}{0pt}{0pt}
    LLM           & 83.7           & 53.2          & 74.6         \\
    WebVid-10M    & 84.2           & 53.8          & 75.5         \\
    Random $\bm{Z}\bm{Z}^\top$        & 75.2           & 46.8          & 69.2         \\
    \cellcolor{backcolor}$\mathcal{N} (0 , I)$     & \cellcolor{backcolor}\textbf{84.9}   & \cellcolor{backcolor}\textbf{54.2}   & \cellcolor{backcolor}\textbf{75.7}      \\
    \end{tabular}%
\end{subtable}%
\end{minipage}
\hfill
\begin{minipage}{0.32\textwidth}
\setlength{\tabcolsep}{0.7mm}
\begin{subtable}{1.0\linewidth}
\centering
    \caption{Various distillation objectives.}
    \label{Tab::AS::OOD}
    \begin{tabular}{c|ccc}
    Type                            & UCF             & HMDB           & K600          \\ 
    \specialrule{0.7pt}{0pt}{0pt}
    L2 Loss                  & 83.7    & 53.1      & 75.1          \\
    KL divergence            & 83.8    & 53.6      & 75.0          \\
    Relation NCE        & 83.9    & 53.0      & 74.7          \\
    \cellcolor{backcolor}$\mathcal{L}_{\mathrm{RSD}}$            
    & \cellcolor{backcolor}\textbf{84.9}    
    & \cellcolor{backcolor}\textbf{54.2}    
    & \cellcolor{backcolor}\textbf{75.7}       \\
    \end{tabular}%
\end{subtable}%
\end{minipage}
\hfill
\begin{minipage}{0.32\textwidth}
\setlength{\tabcolsep}{1.7mm}
\begin{subtable}{1.0\linewidth}
\centering
    \caption{Number of the random geometric anchors $M$ in distillation.}
    \label{Tab::AS::OODnumber}
    \begin{tabular}{c|cccc}
    Number        & UCF            & HMDB          & K600         \\
    \specialrule{0.7pt}{0pt}{0pt}
    2 & 83.9           & 53.0          & 74.7         \\
    20 & 84.6           & 53.9          & 75.4         \\
    \cellcolor{backcolor}200   & \cellcolor{backcolor}\textbf{84.9}           & \cellcolor{backcolor}\textbf{54.2}          & \cellcolor{backcolor}\textbf{75.7}         \\
    2000 & 84.0           & 53.8          & 75.6         \\
    \end{tabular}%
\end{subtable}%
\end{minipage}
\hfill
\begin{minipage}{0.32\textwidth}
\setlength{\tabcolsep}{0.2mm}
\begin{subtable}{1.0\linewidth}
\centering
    \caption{Various implementations of the Specialization Projection.}
    \label{Tab::AS::projection}
    \begin{tabular}{c|cccc}
    Type      & UCF             & K600           & UCF$_f$             & K600$_f$         \\ 
    \specialrule{0.7pt}{0pt}{0pt}
    None           & 82.8             & 73.4          & 81.5         & 69.7\\
    MLP           & 82.2              & 73.3        & \textbf{83.1}    & 71.9      \\
    Transformer           & 81.2              & 72.4         & 82.7    & 71.5      \\
    \cellcolor{backcolor}Two linear          
    & \cellcolor{backcolor}\textbf{83.7}    
    & \cellcolor{backcolor}\textbf{74.8} 
    & \cellcolor{backcolor}82.8
    & \cellcolor{backcolor}\textbf{72.1} \\
    \end{tabular}%
\end{subtable}%
\end{minipage}
\hfill
\begin{minipage}{0.32\textwidth}
\setlength{\tabcolsep}{0.2mm}
\begin{subtable}{1.0\linewidth}
\centering
    \caption{Effects of applying rephrased text and weight ensemble.}
    \label{Tab::AS::Rephrased}
    \begin{tabular}{c|ccc}
    Methods      & UCF             & HMDB           & K600          \\ 
    \specialrule{0.7pt}{0pt}{0pt}
    None           & 84.9             & 54.2  & 75.7          \\
    +Rephrased text           & \textbf{86.0}             & 54.3  & 77.4          \\
    \cellcolor{backcolor}+Weight ensemble           & \cellcolor{backcolor}85.9        & \cellcolor{backcolor}\textbf{54.6}          & \cellcolor{backcolor}\textbf{78.1}          \\
    \multicolumn{4}{c}{}\\
    \end{tabular}%
\end{subtable}%
\end{minipage}
\end{center}
\vspace{-23pt}
\end{table*}

\begin{table}[t]
\vspace{6pt}
\centering
\caption{Structural preservation on unseen categories relative to the original CLIP space. Dir.\ cos.: text-direction cosine; kNN@5: text kNN overlap@5; $\rho$: cross-modal Spearman correlation.}
\label{Tab::UnseenStructure}
\fontsize{8.6pt}{10pt}\selectfont
\setlength{\tabcolsep}{4pt}
\renewcommand{\arraystretch}{0.97}
\definecolor{backcolor}{RGB}{230, 239, 250}
\resizebox{\linewidth}{!}{%
\begin{tabular}{lccc|ccc|ccc}
\toprule
\multirow{2}{*}{Method} & \multicolumn{3}{c}{UCF} & \multicolumn{3}{c}{HMDB} & \multicolumn{3}{c}{K600} \\
\cmidrule(lr){2-4}\cmidrule(lr){5-7}\cmidrule(lr){8-10}
& Dir.\ cos.\ $\uparrow$ & kNN@5 $\uparrow$ & $\rho$ $\uparrow$
& Dir.\ cos.\ $\uparrow$ & kNN@5 $\uparrow$ & $\rho$ $\uparrow$
& Dir.\ cos.\ $\uparrow$ & kNN@5 $\uparrow$ & $\rho$ $\uparrow$ \\
\midrule
Standard fine-tuning & 0.87 & 0.65 & 0.75 & 0.88 & 0.70 & 0.76 & 0.90 & 0.65 & 0.79 \\
\rowcolor{backcolor}
\textbf{TACO (ours)} & \textbf{0.95} & \textbf{0.72} & \textbf{0.84} & \textbf{0.94} & \textbf{0.77} & \textbf{0.83} & \textbf{0.95} & \textbf{0.75} & \textbf{0.84} \\
\bottomrule
\end{tabular}%
}
\end{table}

\subsection{Ablation Studies}
\label{Sec::Experiments::Ablation}
\paragraph{Component-wise analysis of \emph{TACO}.}
To analyze the effect of proposed components, we perform in-depth ablations with the ViT-B/16 network in Table~\ref{Tab::AS::components}.
Our baseline is the ViT-B/16 CLIP model adapted under the standard fine-tuning paradigm.
The results shows that both Specialization Projection and $\mathcal{L}_{\mathrm{RSD}}$ are effective in mitigating harmful alignment shift from two complementary mechanisms: decoupled optimization pathway and structural regularization, respectively. 
Their combination yields a promising synergistic effect, demonstrating the effectiveness of our method.

\paragraph{Construction of the geometric anchors.}
We compare different anchor construction strategies in Table~\ref{Tab::AS::generation}, including semantic anchors derived from LLM-generated video categories~\citep{GPT-4o} and WebVid text descriptions~\citep{WebVid}, with embeddings reduced by K-means clustering~\citep{kmeans}. 
The results show that our Gaussian sampling strategy performs better while requiring no additional text collection or embedding computation. 
This suggests that the gain of $\mathcal{L}_{\mathrm{RSD}}$ does not primarily come from semantic information carried by the anchors, but from providing uniformly distributed geometric references in the shared embedding space. 
Compared with semantically meaningful but distributionally biased anchors, random geometric anchors offer more balanced global coverage. 
We further replace the anchor term in Eq.~\ref{Eq::eq5} with a randomly initialized $D\times D$ matrix, which causes training collapse. 
This shows that the gain does not come from arbitrary randomness. 
Since a random matrix does not lie in the CLIP hypersphere, it introduces unconstrained transformations with no structural meaning. 
Therefore, effective anchors must be valid points in the same representation space as the model.

\paragraph{Effects of regularizing the OOD space.}
In Table~\ref{Tab::AS::OOD}, we compare L2, KL divergence, and a relation-based NCE objective~\cite{CRCD} built from matched relation pairs and queue-based negatives. 
The results show that preserving the structure of the entire representation space is crucial for generalization. 
By regularizing the geometry induced by the constructed OOD anchors, our method better suppresses representation deviation beyond the training distribution and achieves superior performance.

\paragraph{Varying numbers $M$ of random geometric anchors.}
Table~\ref{Tab::AS::OODnumber} shows that increasing $M$ improves results by providing better coverage of the constructed OOD space, and $M=200$ is sufficient to capture the relative structure relationships. 
Further increasing $M$ slightly hurts performance, since the distillation term becomes overly restrictive and hinders task-specific adaptation.

\paragraph{Various implementations of the specialization projection.}
Table~\ref{Tab::AS::projection} compares different specialization projection designs under two fine-tuning settings, depending on whether the text encoder is updated. 
When the text encoder is frozen, all three designs improve performance; however, when it is tunable, MLP and Transformer projections degrade generalization. 
This suggests that the optimization branch should remain close to the original representation space: more expressive projections may over-modify the optimization embedding, especially when the text encoder is also updated, making cross-modal alignment harder to preserve. 
In contrast, two simple linear layers effectively decouple the representation and optimization spaces, yielding consistent gains in both settings.

\paragraph{Effects of the rephrased text and weight ensemble.}
To further enhance generalization, we incorporate the rephrased text descriptions from FROSTER~\citep{FROSTER} and the weight ensemble technique from Open-VCLIP~\citep{OpenVCLIP}. 
As shown in Table~\ref{Tab::AS::Rephrased}, both are orthogonal to TACO and can provide additional gains. 
Notably, TACO already achieves clear improvements without relying on these auxiliary techniques, while rephrased prompts or weight ensembling may further help depending on the training dynamics. 
In particular, we find that weight ensembling is most effective when the model is over-trained, but may be less beneficial otherwise. 

\paragraph{Structural preservation on unseen categories.}
To complement the accuracy comparison, we directly examine structural preservation on real unseen categories using the original CLIP space as a reference. As shown in Table~\ref{Tab::UnseenStructure}, across UCF, HMDB, and K600, TACO consistently improves text-direction cosine, text kNN overlap@5, and cross-modal Spearman correlation over standard fine-tuning. These results provide complementary evidence that TACO better preserves unseen-category directions, local neighborhoods, and cross-modal rankings. Metric definitions are provided in Appendix~\ref{Appendix::Sec::Overlap}.
\emph{Please see supplementary for more ablations.}

\begin{table*}[t]
\renewcommand\arraystretch{0.9}
\caption{Performance comparison with state-of-the-art methods under the base-to-novel setting. “HM” indicates the harmonic mean of the accuracy on base and novel sets.}
\label{Tab::B2N}
\vspace{-2.5ex}
\definecolor{backcolor}{RGB}{230, 239, 250}
\begin{center}
\begin{large}
\resizebox{1.0\linewidth}{!}{
\begin{tabular}{lcccc|ccc|ccc|ccc}
\toprule
\multirow{2}{*}{Method}  & \multirow{2}{*}{Venue}    & \multicolumn{3}{c}{K400} & \multicolumn{3}{c}{HMDB}    & \multicolumn{3}{c}{UCF}    & \multicolumn{3}{c}{SSv2}       \\[-0.6ex] \cmidrule(lr){3-5} \cmidrule(lr){6-8} \cmidrule(lr){9-11} \cmidrule(lr){12-14}\\[-2.8ex]
                        &    & BASE   & Novel & HM    & BASE   & Novel & HM    & BASE  & Novel  & HM    & BASE  & Novel & HM \\[-0.4ex]
\midrule
CLIP~\citep{CLIP}    & ICML'21 & 62.3  & 53.4  & 57.5  & 53.3   & 46.8  & 49.8  & 78.5  & 63.6   & 70.3  & 4.9   & 5.3   & 5.1  \\
ActionCLIP~\citep{ActionClip}  & arXiv'21
                              & 61.0  & 46.2  & 52.6  & 69.1   & 37.3  & 48.5  & 90.1  & 58.1   & 70.7  & 13.3  & 10.1  & 11.5 \\
X-CLIP~\citep{XCLIP} & ECCV'22 & 74.1  & 56.4  & 64.0  & 69.4   & 45.5  & 55.0  & 89.9  & 58.9   & 71.2  & 8.5   & 6.6   & 7.4  \\
VPT~\citep{VPT}      & ECCV'22 & 69.7  & 37.6  & 48.8  & 46.2   & 16.0  & 23.8  & 90.5  & 40.4   & 55.8  & 8.3   & 5.3   & 6.4  \\
AIM~\citep{AIM}     & ICLR'23 & 74.6  & 62.5  & 68.0  & 64.0   & 51.6  & 57.1  & 89.8  & 76.4   & 82.6  & 8.5   & 7.9   & 8.2  \\
ST-Adapter~\citep{ST-A}       & NeurIPS'22 
                              & 73.6  & 62.0  & 67.3  & 65.3   & 48.9  & 55.9  & 85.5  & 76.8   & 80.9  & 9.3   & 8.4   & 8.8  \\
ViFi-CLIP~\citep{ViFiCLIP}    & CVPR'23
                              & 76.4  & 61.1  & 67.9  & 73.8   & 53.3  & 61.9  & 92.9  & 67.7   & 78.3  & 16.2  & 12.1  & 13.9  \\
Open-VCLIP~\citep{OpenVCLIP}  & ICML'23
                              & 76.5  & 62.6  & 68.9  & 70.3   & 50.4  & 58.7  & 94.8  & 77.5   & 85.3  & 16.0  & 11.0  & 13.0  \\
FROSTER~\citep{FROSTER}       & ICLR'24 
                              & \underline{77.8}  & \textbf{64.3}  & \textbf{70.4}  & \underline{74.1}   & \underline{58.0}  & \underline{65.1}  & \underline{95.3}  & \underline{80.0}   & \underline{87.0}  & \textbf{18.3}  & 12.2  & \textbf{14.6}  \\
Open-MeDe~\citep{OpenMeDe}    & ICCV'25 
                              & 77.2  & \underline{63.8}  & 69.9  & 73.6   & 56.4  & 63.9  & 94.9  & 78.5   & 85.9  & 17.1  & \underline{12.3}  & \underline{14.3}  \\
\cellcolor{backcolor}\textbf{TACO (ours)} &
\cellcolor{backcolor} &
\cellcolor{backcolor}\textbf{78.2} & 
\cellcolor{backcolor}\underline{63.8} & 
\cellcolor{backcolor}\underline{70.3} & 
\cellcolor{backcolor}\textbf{74.4} &  
\cellcolor{backcolor}\textbf{62.3} & 
\cellcolor{backcolor}\textbf{67.8} & 
\cellcolor{backcolor}\textbf{95.8} & 
\cellcolor{backcolor}\textbf{82.2} & 
\cellcolor{backcolor}\textbf{88.5} &  
\cellcolor{backcolor}\underline{17.7} & 
\cellcolor{backcolor}\textbf{12.4} & 
\cellcolor{backcolor}\textbf{14.6} \\
\bottomrule
\end{tabular}
}
\end{large}
\end{center}
\vskip -0.7cm
\end{table*}

\subsection{Main Results}
\label{Sec::Experiments::MR}

\paragraph{Base-to-novel video recognition.}
Table~\ref{Tab::B2N} reports the results under the base-to-novel setting, which reflects the model’s joint ability to fit video-specific biases while being adapted to unknown categories. 
TACO achieves the best or competitive harmonic-mean performance across all four benchmarks, with especially clear gains on HMDB and UCF. 
Notably, the improvements are mainly driven by better recognition of novel classes, suggesting that our adaptation does not simply bias the model toward the training categories. 
This observation is consistent with our method design: by preserving the broader geometric structure of the pretrained space, TACO enables video adaptation while better maintaining the alignment needed for unseen categories.
The gains on SSv2 are more modest, likely because SSv2 requires stronger temporal reasoning, whereas our method mainly targets task-consistent open-vocabulary adaptation. Still, TACO remains competitive without introducing additional temporal modules, indicating that it is complementary to dedicated temporal modeling techniques.

\paragraph{Cross-dataset video recognition.}
Table~\ref{Tab::Zeroshot} presents comparisons with the state-of-the-art methods under the cross-dataset setting, which assesses the model's generalization towards out-of-distribution categories.
Under this protocol, TACO consistently outperforms prior methods across all datasets and both backbone scales, and the same advantage is retained on the full-dataset evaluation in Table~\ref{Tab::ZeroshotFull}.
These gains are consistent with Figure~\ref{Fig::Relativesim} and Figure~\ref{Fig::AlignmentShift}: TACO reduces representation drift in OOD regions and maintains a more moderate visual-text alignment shift during adaptation, both of which are important for open-vocabulary generalization. This matches our design, where $\mathcal{L}_{\mathrm{RSD}}$ regularizes relations beyond the training categories and the Specialization Projection reduces over-specialization of the representation space. As a result, TACO better preserves the transferable structure inherited from CLIP, especially under larger distribution shifts such as HMDB and Kinetics-600.

\paragraph{Applicability to various adaptation methods.}
Table~\ref{Tab::Network} further shows that TACO can be integrated with adapter-based, prompt-based, LoRA-based, and fully fine-tuned adaptation methods, consistently improving their zero-shot generalization. 
This suggests that our framework is largely orthogonal to the underlying adaptation mechanism, since it acts by preserving representation geometry and decoupling it from the optimization space.
The strongest absolute performance is obtained with full fine-tuning, suggesting that broader parameter updates provide greater capacity for video adaptation, while our method helps prevent this flexibility from causing excessive drift in the pretrained space.

\begin{table*}[t]
\renewcommand\arraystretch{0.9}
\setlength{\tabcolsep}{11pt}
\caption{Zero-shot classification performance compared with the state-of-the-art methods under the cross-dataset setting, evaluated on the \emph{validation splits} of UCF-101, HMDB-51, and Kinetics-600.  }
\vspace{-2.5ex}
\label{Tab::Zeroshot}
\definecolor{backcolor}{RGB}{230, 239, 250}
\begin{center}
\begin{normalsize}
\resizebox{1.0\linewidth}{!}{
\begin{tabular}{lcccccc}
\toprule
Method                         & Venue     & Encoder   & Frames & UCF-101       & HMDB-51      & Kinetics-600 \\
\midrule
ActionCLIP~\citep{ActionClip}   & arXiv'21  & ViT-B/16  & 32     & 58.3$\pm$3.4  & 40.8$\pm$5.4 & 67.7$\pm$1.1 \\
X-CLIP~\citep{XCLIP}            & ECCV'22   & ViT-B/16  & 32     & 72.0$\pm$2.3  & 44.6$\pm$5.2 & 65.2$\pm$0.4 \\
ST-Adapter~\citep{ST-A}         & NeurIPS'22& ViT-B/16  & 8      & 76.9$\pm$0.8  & 51.5$\pm$0.6 & 60.2$\pm$1.8 \\
Vita-CLIP~\citep{VitaCLIP}      & CVPR'23   & ViT-B/16  & 8/32   & 75.0$\pm$0.6  & 48.6$\pm$0.6 & 67.4$\pm$0.5 \\
ViFi-CLIP~\citep{ViFiCLIP}      & CVPR'23   & ViT-B/16  & 32     & 76.8$\pm$0.7  & 51.3$\pm$0.6 & 71.2$\pm$1.0 \\
Open-VCLIP~\citep{OpenVCLIP}    & ICML'23   & ViT-B/16  & 8      & 83.4$\pm$1.2  & 53.9$\pm$1.2 & 73.0$\pm$0.8 \\
MAXI~\citep{MAXI}               & ICCV'23   & ViT-B/16  & 16/32  & 78.2$\pm$0.8  & 52.3$\pm$0.7 & 71.5$\pm$0.8 \\
FROSTER~\citep{FROSTER}         & ICLR'24   & ViT-B/16  & 8      & \underline{84.8$\pm$1.1}  
                                                                                 & 54.8$\pm$1.3 & \underline{74.8$\pm$0.9} \\
OST~\citep{OST}                 & CVPR'24   & ViT-B/16  & 8      & 77.9$\pm$1.3  & 54.9$\pm$1.1 & 73.9$\pm$0.8 \\
MoTE~\citep{MoTE}            & NeurIPS'24& ViT-B/16  & 8      & 83.4$\pm$0.7  & \underline{55.8$\pm$0.9} 
                                                                                                & 70.2$\pm$0.6 \\
Open-MeDe~\citep{OpenMeDe}      & ICCV'25   & ViT-B/16  & 8      & 83.7$\pm$1.3  & 54.6$\pm$1.1 & 73.7$\pm$0.9 \\
\cellcolor{backcolor}\textbf{TACO (ours)} &
\cellcolor{backcolor} & 
\cellcolor{backcolor}ViT-B/16  & 
\cellcolor{backcolor}8      & 
\cellcolor{backcolor}\textbf{85.6}$\pm$1.2  & 
\cellcolor{backcolor}\textbf{60.0}$\pm$0.5    &  
\cellcolor{backcolor}\textbf{77.0}$\pm$0.9 \\
\midrule
X-Florence~\citep{XCLIP}        & ECCV'22   & Florence  & 32     & 73.2$\pm$4.2  & 48.4$\pm$4.9 & 68.8$\pm$0.9 \\
Text4Vis~\citep{Text4vis}       & AAAI'23   & ViT-L/14  & 8      & 82.6$\pm$0.7  & 52.4$\pm$0.4 & 72.1$\pm$0.9 \\
OTI~\citep{OTI}                 & ACMMM'23  & ViT-L/14  & 8      & 88.1$\pm$1.0  & 59.3$\pm$1.7 & 70.6$\pm$0.5 \\
Open-VCLIP~\citep{OpenVCLIP}    & ICML'23   & ViT-L/14  & 8      & 87.6$\pm$1.2  & 59.0$\pm$0.6 & \underline{81.1}$\pm$0.8 \\
DiST~\citep{DiST}               & ICCV'23   & ViT-L/14  & 32     & 74.9$\pm$0.8  & 57.5$\pm$1.6 & 75.0$\pm$0.7 \\
MoTE~\citep{MoTE}            & NeurIPS'24& ViT-L/14  & 8      & \underline{88.7$\pm$0.6}  & \underline{61.4$\pm$1.3} & 78.4$\pm$0.9 \\
\cellcolor{backcolor}\textbf{TACO (ours)}    &
\cellcolor{backcolor}    & 
\cellcolor{backcolor}ViT-L/14  & 
\cellcolor{backcolor}8      & 
\cellcolor{backcolor}\textbf{91.4}$\pm$0.7  & 
\cellcolor{backcolor}\textbf{64.2}$\pm$0.8 & 
\cellcolor{backcolor}\textbf{83.9}$\pm$0.7 \\
\bottomrule
\end{tabular}
}
\end{normalsize}
\end{center}
\vskip -0.7cm
\end{table*}

\begin{table}[t]
\begin{minipage}{0.49\linewidth}
    \renewcommand\arraystretch{0.92}
    \setlength{\tabcolsep}{8pt}
    \caption{Zero-shot performance of UCF and HMDB on the \emph{full} dataset.$\,$* denotes evaluation with the full validation set on HMDB.}
    \label{Tab::ZeroshotFull}
    \vspace{-1ex}
    \definecolor{backcolor}{RGB}{230, 239, 250}
    \begin{center}
    \begin{normalsize}
    \resizebox{\linewidth}{!}{
    \begin{tabular}{lccc}
    \toprule
    Method                          & Encoder       & UCF       & HMDB  \\
    \midrule
    CLIP~\citep{CLIP}               & ViT-B/16      & 74.9      & 46.7  \\
    AIM~\citep{AIM}                 & ViT-B/16      & 79.0      & 49.5  \\
    ST-Adapter~\citep{ST-A}         & ViT-B/16      & 77.9      & 50.3  \\
    Open-VCLIP*~\citep{OpenVCLIP}    & ViT-B/16      & 83.5      & 53.2  \\
    FROSTER*~\citep{FROSTER}         & ViT-B/16      & \underline{85.0}      
                                                    & \underline{54.5}  \\
    \cellcolor{backcolor}\textbf{TACO (ours)} &
    \cellcolor{backcolor} ViT-B/16 &
    \cellcolor{backcolor} \textbf{85.9} &
    \cellcolor{backcolor} \textbf{54.6} \\
    \midrule
    Text4Vis~\citep{Text4vis}       & ViT-L/14      & 79.6      & 49.8  \\
    BIKE~\citep{Text4vis}           & ViT-L/14      & 80.8      & 52.8  \\
    OTI~\citep{OTI}                 & ViT-L/14      & 88.3      & 55.8  \\
    MoTE~\citep{MoTE}               & ViT-L/14      & \underline{89.4}      
                                                    & \underline{56.3}  \\
    \cellcolor{backcolor}\textbf{TACO (ours)} &
    \cellcolor{backcolor} ViT-L/14 &
    \cellcolor{backcolor} \textbf{91.6} &
    \cellcolor{backcolor} \textbf{59.9} \\
    \bottomrule
    \end{tabular}
    }
    \end{normalsize}
    \end{center}
\end{minipage}
\hfill
\begin{minipage}{0.49\linewidth}
    \renewcommand\arraystretch{0.92}
    \caption{Integrating our TACO with various adaptation methods can significantly improve the zero-shot generalization.}
    \label{Tab::Network}
    \vspace{-1ex}
    \definecolor{backcolor}{RGB}{230, 239, 250}
    \begin{center}
    \begin{normalsize}
    \resizebox{1.0\linewidth}{!}{
    \begin{tabular}{l|c|ccc}
    \toprule
        Type    & Method                    & UCF           & HMDB          & K600  \\
    \midrule
    \multirow{3}{*}{Adapter-based} 
                & ST-Adapter~\citep{ST-A}   & 77.9          & 50.3          & 60.2  \\
                & \cellcolor{backcolor} + \emph{TACO(ours)}       
                & \cellcolor{backcolor}\textbf{80.5}    
                & \cellcolor{backcolor}\textbf{53.1} 
                & \cellcolor{backcolor}\textbf{73.9} \\
                & $\Delta$                  
                & \textbf{+2.6}         
                & \textbf{+2.8}         
                & \textbf{+13.7} \\
    \midrule
    \multirow{3}{*}{Prompt-based} 
                & Vita-CLIP~\citep{VitaCLIP}& 78.6          & 50.5          & 67.4   \\
                & \cellcolor{backcolor} + \emph{TACO(ours)}       
                & \cellcolor{backcolor}\textbf{80.3}    
                & \cellcolor{backcolor}\textbf{52.6} 
                & \cellcolor{backcolor}\textbf{73.2} \\
                & $\Delta$                  
                & \textbf{+1.7}         
                & \textbf{+2.1}         
                & \textbf{+5.8} \\
    \midrule
    \multirow{3}{*}{LoRA-based} 
                & CLIP + LoRA~\citep{LoRA}  & 80.5          & 50.4          & 71.2   \\
                & \cellcolor{backcolor} + \emph{TACO(ours)}       
                & \cellcolor{backcolor}\textbf{83.1}    
                & \cellcolor{backcolor}\textbf{54.4} 
                & \cellcolor{backcolor}\textbf{74.7} \\
                & $\Delta$                  
                & \textbf{+2.6}         
                & \textbf{+4.0}         
                & \textbf{+3.5} \\
    \midrule
    \multirow{3}{*}{Fully-tuned} 
                & Fine-tuned CLIP & 82.8          & 52.0          & 73.4   \\
                & \cellcolor{backcolor} + \emph{TACO(ours)}       
                & \cellcolor{backcolor}\textbf{84.9}    
                & \cellcolor{backcolor}\textbf{54.2} 
                & \cellcolor{backcolor}\textbf{75.7} \\
                & $\Delta$                  
                & \textbf{+2.1}         
                & \textbf{+2.2}         
                & \textbf{+2.3} \\
    \bottomrule
    \end{tabular}
    }
    \end{normalsize}
    \end{center}
\end{minipage}
\vskip -0.5cm
\end{table}

\section{Visualizations}
\label{Sec::Visualizations}

\begin{figure}[t]
\centering
\includegraphics[width=\linewidth]{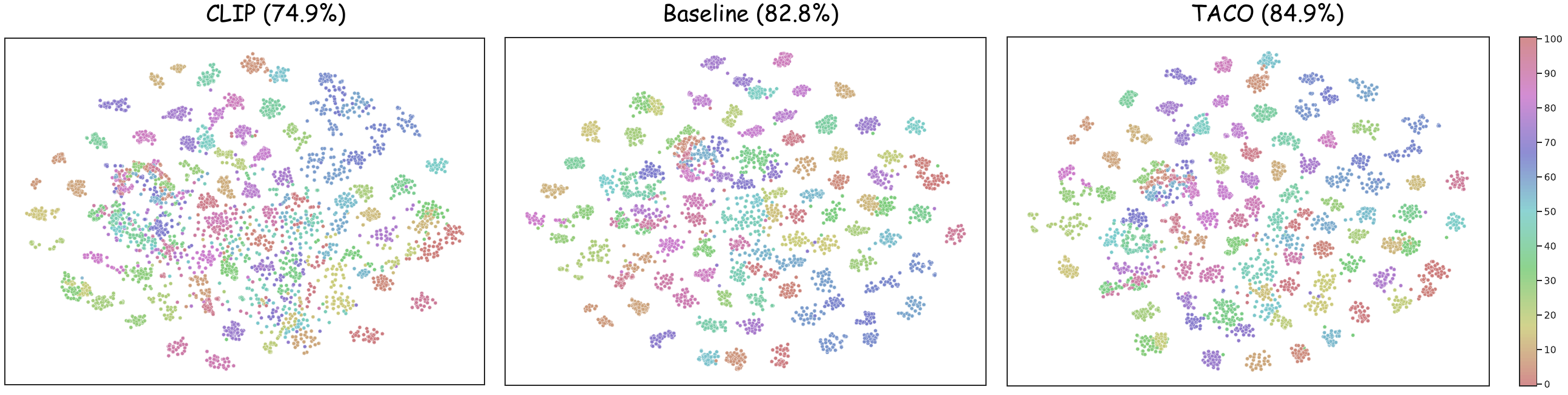}
\vspace{-4pt}
\caption{t-SNE~\cite{tsne} visualization of UCF-101 visual representations from CLIP, standard fine-tuning, and TACO (left to right).}
\label{Fig::tsne_main}
\vspace{-6pt}
\end{figure}

To obtain a more intuitive understanding of the representation distribution, we present t-SNE~\cite{tsne} visualizations of the visual representations from CLIP, Fine-tuned CLIP, and our model in Figure~\ref{Fig::tsne_main}.
The representations are extracted from the UCF-101 dataset.
As shown in the figure, the representation distribution of the CLIP model is relatively messy due to a lack of video-specific knowledge.
Compared with the standard fine-tuned model, our TACO exhibits a more compact intra-class distribution and more pronounced inter-class distinctions.
This phenomenon demonstrates that our approach can introduce new video knowledge while preserving the relative structure among unknown categories, thereby achieving greater performance improvements, which can be attributed to the introduction of Relative Structure Distillation.

\section{Related Work}
\label{Sec::RelatedWork}


\paragraph{Adapting VLMs to video recognition.}
Adapting VLMs to video recognition has proven effective, but remains challenging in balancing video-specific adaptation with the pretrained generalization of VLMs~\citep{ViFiCLIP}. 
To better capture temporal dynamics, existing methods introduce additional parameterized modules, such as adapters~\citep{ST-A, AIM}, prompts~\citep{A5,VitaCLIP, VPT}, and Transformer layers~\citep{Text4vis,BIKE,MoTE}. 
For example, X-CLIP~\citep{XCLIP} proposes an attention-based multi-frame integration module for cross-frame information exchange. 
Another line of work focuses on preserving generalization during adaptation: Open-VCLIP~\citep{OpenVCLIP} interpolates model weights along the optimization trajectory, FROSTER~\cite{FROSTER} alleviates the overfitting by ensuring the learned features do not diverge too far from the frozen CLIP, and Open-MeDe~\citep{OpenMeDe} uses meta-optimization to mitigate the static bias of the pretrained model. 
Despite their strong open-vocabulary performance, these methods still optimize mainly within the training distribution, leaving the mismatch between adaptation and open-vocabulary evaluation insufficiently addressed. 
Our method revisits this issue from the perspective of representation consistency beyond the training distribution and achieves stronger generalization.


\paragraph{Knowledge distillation for VLMs.}
Knowledge distillation has been widely used to adapt pretrained models and improve generalization~\citep{Clipping, Promptkd}. 
Its core idea is to transfer the teacher’s knowledge by matching logits, features, or intermediate representations, either to improve the student’s generalization~\citep{CLIP-KD, KDPL, VLKD} or constrain fine-tuning with pretrained supervision~\citep{FROSTER, VL2V}. 
For example, CLIPPING~\citep{Clipping} distills layer-wise knowledge from a larger teacher into an efficient student model, while FROSTER~\citep{FROSTER} reduces overfitting through residual feature distillation. 
Relation-based distillation methods such as CRCD~\cite{CRCD} are also related, but they are designed for closed-set classification and transfer local relations among training samples, whereas our method uses random geometric anchors to regularize the global geometry of the entire embedding space for open-vocabulary adaptation. 
Overall, existing distillation methods mainly operate within the training distribution. 
In contrast, we emphasize preserving representation consistency beyond the training distribution and propose a distillation objective that maintains the relative structure of the OOD space during adaptation.
\section{Conclusion}
\label{Sec::Conclusion}
In this work, we present \emph{TACO}, a concise adaptation framework to mitigate the detrimental effects induced by representation deviations beyond the training distribution.
By analyzing the objective inconsistency of existing adaptation paradigms, we formulate a concrete adaptation principle and propose \emph{Relative Structure Distillation} to enforce consistent relative structures across the entire representation space. 
Furthermore, we introduce a space decoupling strategy with a specialization projection to alleviate the overspecialization.
Extensive experiments demonstrate that \emph{TACO} achieves state-of-the-art results across diverse benchmarks under cross-dataset and base-to-novel settings.

\clearpage
\section*{Acknowledgements}
This paper is supported by the National Natural Science Foundation of China (Grant No. 62333017).


{\small
\bibliographystyle{plainnat}
\bibliography{References}
}

\newpage
\appendix
\onecolumn

\section*{Appendix Overview}
This appendix is organized as follows:
\begin{itemize}
    \item \textbf{Section~\ref{Appendix::Sec::Limitation}}: Limitations and Broader Impact.
    \item \textbf{Section~\ref{Appendix::Sec::details}}: More Implementation Details.
    \item \textbf{Section~\ref{Appendix::Sec::Ablations}}: More Ablations.
    \item \textbf{Section~\ref{Appendix::Sec::Theory}}: Theoretical Analysis of Relative Structure Distillation.
    \item \textbf{Section~\ref{Appendix::Sec::Overlap}}: More Analysis and Discussions.
    \item \textbf{Section~\ref{Appendix::Sec::Visualization}}: Visualizations.
    \item \textbf{Section~\ref{Appendix::Sec::TPT}}: Textual Prompts Used in Evaluation.
    \item \textbf{Section~\ref{Appendix::Sec::Dataset}}: Dataset Details.
\end{itemize}



\section{Limitations and Broader Impact}
\label{Appendix::Sec::Limitation}
\paragraph{Limitations}
While our approach achieves strong performance on open-vocabulary tasks, we observe that its improvements are relatively limited on datasets where temporal information plays a critical role.  A promising direction for future research is to integrate our framework with existing temporal modeling techniques to better handle such scenarios.
In addition, our method requires setting a relatively small momentum parameter for the teacher model to maintain consistency in the OOD space. However, we observe that the optimal value of this parameter may vary across datasets, which could limit the robustness of our approach. A potential future direction is to develop a more stable strategy for updating the teacher model.

\paragraph{Broader Impact}
Adapting foundation models to downstream tasks has become a prevailing trend in machine learning~\cite{cocoop, FROSTER, groundvts, mlct}. 
We argue that exploring effective strategies for adapting vision-language models to open-vocabulary tasks is both timely and necessary for real-world applications. 
This work seeks to offer insights that contribute to the broader and long-term use of foundation models. 
Moreover, our research provides new perspectives in understanding the adaptation of VLMs in terms of maintaining the representation alignment beyond the training distribution.
We believe this insight can contribute to a deeper understanding of existing adaptation strategies (e.g., weight ensemble~\cite{OpenVCLIP}) and the development of new methods.
Besides, while our study focuses on video recognition, which has wide-ranging applications such as surveillance, it is crucial that concerns regarding privacy and individual rights are thoroughly considered prior to practical deployment.

\begin{table}[ht]
\renewcommand\arraystretch{1.05}
\caption{Hyper-parameter details during fine-tuning.}
\definecolor{midgrey}{RGB}{225,225,225}
\centering
\scalebox{0.95}{
\begin{tabular}{l|c} \toprule
    ~ & Value \\ \midrule
    \multicolumn{2}{l}{\cellcolor{midgrey}\emph{Optimization details}} \\
    Batch size & 192 \\
    Optimizer & AdamW \\
    Weight decay & 0.2 \\
    Adam $\beta_1$,$\beta_2$ & 0.9, 0.999 \\ 
    Learning rate (Projection) & 5e-5 \\
    Learning rate (CLIP layers) & 3.75e-6  \\
    Learning rate decay         & Cosine \\
    Training epochs & 15 \\
    Linear warm-up epochs & 5 (cross-dataset), 2 (base-to-novel) \\
    \midrule
    \multicolumn{2}{l}{\cellcolor{midgrey}\emph{Augmentation}} \\
    RandomResizedCrop   &         \\
    \quad\quad\quad Area          &    [0.08, 1.00] \\
    \quad\quad\quad Aspect ratio  &    [3/4 , 4/3]  \\
    \quad\quad\quad Crop size     &    224  \\
    Random Horizontal Flip & 0.5 \\
    Random Gray scale & 0.2 \\
    \bottomrule
\end{tabular}}
\label{Appendix::Tab::para}
\end{table}

\section{More Implementation Details}
\label{Appendix::Sec::details}
In Table \ref{Appendix::Tab::para}, we present the hyper-parameters set for optimization.
For both the Cross-dataset and Base-to-novel settings, we trained the model for 15 epochs.
All experiments are conducted using 3 NVIDIA GeForce RTX 4090.

For cross-dataset evaluation, the methods are evaluated on three official splits or the full dataset of UCF-101 and HMDB-51. For Kinetics-600, we adopt the three splits provided by \citep{ER}. 
Each split contains 160 categories out of 220 new categories that do not exist in K400.
We report the average Top-1 accuracy and the standard deviation on three splits.
To further enhance the generalization performance, we incorporate the rephrased text descriptions from FROSTER~\citep{FROSTER} and the weight ensemble technique~\citep{OpenVCLIP} into our method.

For base-to-novel evaluation, we do not apply weight ensemble since it may affect the model's performance on the base category. 
Following the previous work~\citep{FROSTER}, we employed the rephrased text descriptions in this setting.
Besides, we do not apply the rephrased text descriptions to the SSv2 dataset.

\section{More Ablations}
\label{Appendix::Sec::Ablations}
\paragraph{Training and efficiency analysis of TACO.}
\begin{wraptable}[6]{r}{0.38\textwidth}
    \renewcommand\arraystretch{1.0}
\vspace{-2ex}
\setlength{\tabcolsep}{0.5pt}
\definecolor{backcolor}{RGB}{230, 239, 250}
\caption{Training and efficiency analysis of TACO.}
\label{Appendix::Tab::Ablation::Cost}
\vspace{-3ex}
\begin{center}
\begin{normalsize}
\resizebox{1.0\linewidth}{!}{
\begin{tabular}{l|cccc}
\toprule
Method & GPU-hours & GFLOPs & Mem (G) & Param (M) \\
\midrule
Baseline & 36.3 & 432 & 12.0 & 86.2 \\
+ spec. proj. & 36.4 & 432 & 12.1 & 86.2 \\
\cellcolor{backcolor}+ $\mathcal{L}_{\mathrm{RSD}}$ & \cellcolor{backcolor}37.8 & \cellcolor{backcolor}576 & \cellcolor{backcolor}15.2 & \cellcolor{backcolor}86.2  \\
\bottomrule
\end{tabular}
}
\end{normalsize}
\end{center}
\vskip -0.5cm
\end{wraptable}
We report the training and inference cost of different components in Table~\ref{Appendix::Tab::Ablation::Cost}. Training time is measured on 3 NVIDIA GeForce RTX 4090 GPUs with a batch size of 192, and GPU-hours are computed from the wall-clock time. We also report GFLOPs, memory usage, parameter count, and throughput (TP) under the same ViT-B/16 setting.
As shown in the table, the specialization projection introduces almost no extra overhead over the baseline, with nearly unchanged GFLOPs, memory, parameter count, and throughput. Adding $\mathcal{L}_{\mathrm{RSD}}$ increases training cost and memory/FLOPs due to the extra teacher forward pass and anchor-based distillation, but the overall overhead remains modest. These results show that TACO improves performance with limited additional cost.
\paragraph{Effects of varying momentum coefficients $\alpha$.}
\begin{wraptable}[8]{r}{0.38\textwidth}
\vspace{-0.4cm}
\renewcommand\arraystretch{0.6}
\setlength{\tabcolsep}{16pt}
\definecolor{backcolor}{RGB}{230, 239, 250}
\caption{Effects of EMA update mechanism and varying momentum coefficients.}
\label{Appendix::Tab::Ablation::momentum}
\vspace{-2.5ex}
\begin{center}
\begin{normalsize}
\resizebox{1.0\linewidth}{!}{
\begin{tabular}{l|ccc}
     \toprule
            Momentum                          & UCF & HMDB   \\
            \midrule
            1.0             & 84.2 & 53.7    \\
            0.99998         & 84.2 & 54.0        \\
            \cellcolor{backcolor}0.9998          & \cellcolor{backcolor}\textbf{84.9} & \cellcolor{backcolor}\textbf{54.2}  \\
            0.998           & 84.1 & 53.8       \\
            \bottomrule
\end{tabular}
}
\end{normalsize}
\end{center}
\vskip -0.5cm
\end{wraptable}
We further study the impact of the teacher update mechanism and the momentum coefficient on generalization performance. As shown in Table~\ref{Appendix::Tab::Ablation::momentum}, using a frozen teacher model (i.e., momentum $=1.0$) yields inferior results, suggesting that an entirely static teacher may overly constrain the student and hinder the acquisition of new task-specific knowledge. In contrast, the EMA update strategy provides a better balance between preserving structural consistency and gradually incorporating newly learned knowledge. Among the tested values, $\alpha=0.9998$ gives the best performance on both UCF and HMDB. Using either a larger or smaller momentum leads to slight degradation, indicating that an appropriate update rate is important for maintaining the consistency of the OOD space structure during adaptation.


\paragraph{Comparison with conventional regularizations.}
\begin{wraptable}[8]{r}{0.38\textwidth}
\vspace{-0.4cm}
\renewcommand\arraystretch{1.0}
\setlength{\tabcolsep}{10pt}
\definecolor{backcolor}{RGB}{230, 239, 250}
\caption{Comparison with conventional regularization strategies.}
\label{Appendix::Tab::Ablation::regularization}
\vspace{-2.5ex}
\begin{center}
\begin{normalsize}
\resizebox{1.0\linewidth}{!}{
\begin{tabular}{l|cc}
     \toprule
            Method          & UCF & HMDB \\
            \midrule
            Dropout         & 82.9 & 51.6  \\
            Weight decay    & 82.8 & 52.0  \\
            \cellcolor{backcolor}Spec. Proj. & \cellcolor{backcolor}\textbf{83.7} & \cellcolor{backcolor}\textbf{53.1} \\
            \bottomrule
\end{tabular}
}
\end{normalsize}
\end{center}
\vskip -0.5cm
\end{wraptable}
We further compare our Specialization Projection with conventional regularization strategies, including dropout and weight decay. As shown in Table~\ref{Appendix::Tab::Ablation::regularization}, both dropout and weight decay provide only limited improvements in generalization performance. In contrast, specialization projection consistently achieves better results on all three benchmarks, with clear gains on UCF, HMDB, and K600. This suggests that the benefit of our design does not mainly come from generic overfitting reduction. Rather, by explicitly decoupling the representation space from the optimization space, our strategy better prevents the shared visual representation from being directly overspecialized to the fine-tuning categories.

\newpage
\paragraph{Effects of adding the specialization projection during fine-tuning and evaluation.}
\begin{wraptable}[8]{r}{0.38\textwidth}
\vspace{-0.0cm}
\renewcommand\arraystretch{0.9}
\setlength{\tabcolsep}{8pt}
\definecolor{backcolor}{RGB}{230, 239, 250}
\caption{Effects of adding the specialization projection during fine-tuning and evaluation.}
\label{Appendix::Tab::Ablation::speval}
\vspace{-1.5ex}
\begin{center}
\begin{normalsize}
\resizebox{1.0\linewidth}{!}{
\begin{tabular}{cc|cc}
     \toprule
            Fine-tuning     & Evaluation    & UCF & HMDB     \\
            \midrule
                -        &         -     & 82.8 & 52.0    \\
            \cellcolor{backcolor}\ding{51}       &     \cellcolor{backcolor}-         & \cellcolor{backcolor}\textbf{83.7} & \cellcolor{backcolor}\textbf{53.2}    \\
            \ding{51}       & \ding{51}     & 83.3 & 52.2    \\

            \bottomrule
\end{tabular}
}
\end{normalsize}
\end{center}
\vskip -0.5cm
\end{wraptable}
In this ablation study, we further investigate the effect of adding the specialization projection on the generalization of representations during training and evaluation. 
Adding SP during evaluation indicates that the projected representation is used.
As shown in Table~\ref{Appendix::Tab::Ablation::speval}, decoupling the representation space facilitates enhancing model generalization.
Besides, discarding the projection layer during evaluation is important, as the projected representation is more prone to overfitting the fine-tuning data distribution.



\paragraph{Effectiveness of each proposed component on the fine-tuned model with the frozen text encoder.}
\begin{wraptable}[7]{r}{0.38\textwidth}
\vspace{-0.4cm}
\renewcommand\arraystretch{0.9}
\setlength{\tabcolsep}{6pt}
\definecolor{backcolor}{RGB}{230, 239, 250}
\caption{Effectiveness of each proposed component on the fine-tuned model with the frozen text encoder.}
\label{Appendix::Tab::Ablation::components_frozen}
\vspace{-1.5ex}
\begin{center}
\begin{normalsize}
\resizebox{1.0\linewidth}{!}{
\begin{tabular}{cc|ccc}
\toprule
     Spec. Proj.     & $\mathcal{L}_{\mathrm{RSD}}$   &  UCF$_{f}$      & HMDB$_{f}$       & K600$_{f}$    \\
    \specialrule{0.7pt}{0pt}{0pt}
                    &                       & 81.5       & 50.4       & 69.7   \\    
    \ding{51}       &                       & 82.8       & 52.7       & 72.1   \\ 
                    & \ding{51}             & 83.2       & 52.2       & 72.3   \\
    \cellcolor{backcolor}\ding{51}       & \cellcolor{backcolor}\ding{51}             & \cellcolor{backcolor}\textbf{83.8}       & \cellcolor{backcolor}\textbf{53.6}       & \cellcolor{backcolor}\textbf{73.1}   \\
\bottomrule
\end{tabular}
}
\end{normalsize}
\end{center}
\vskip -0.5cm
\end{wraptable}
In Table~\ref{Appendix::Tab::Ablation::components_frozen}, we study the effectiveness of each proposed component when freezing the text encoder during fine-tuning, which is a common design choice in VLMs adaptation. 
Our proposed components deliver significant improvements in generalization performance, demonstrating the versatility of our method under diverse fine-tuning scenarios.


\paragraph{Effects of varying loss balance coefficients $\lambda$ for relative structure distillation.}
\begin{wraptable}[6]{r}{0.38\textwidth}
\vspace{-0.4cm}
\renewcommand\arraystretch{0.8}
\setlength{\tabcolsep}{22pt}
\definecolor{backcolor}{RGB}{230, 239, 250}
\caption{Effects of varying loss balance coefficients.}
\label{Appendix::Tab::Ablation::lambda}
\vspace{-1.5ex}
\begin{center}
\begin{normalsize}
\resizebox{1.0\linewidth}{!}{
\begin{tabular}{c|cc}
\toprule
$\lambda$   &  UCF      & HMDB      \\
    \specialrule{0.7pt}{0pt}{0pt}
0.0         & 83.7       & 53.2     \\  
\cellcolor{backcolor}0.4         & \cellcolor{backcolor}\textbf{84.9}       & \cellcolor{backcolor}54.2     \\    
0.7         & 84.6       & 54.2     \\  
1.0         & 84.5       & \textbf{54.4}     \\    
\bottomrule
\end{tabular}
}
\end{normalsize}
\end{center}
\vskip -0.5cm
\end{wraptable}
In Table~\ref{Appendix::Tab::Ablation::lambda}, we investigate the effects of varying loss balance coefficients $\lambda$ in fine-tuning. 
As shown in the table, incorporating our relative structure distillation objective effectively enhances the generalization performance.
A relatively small loss balance coefficient of 0.4 yields the best overall results.

\paragraph{Effects of varying learning rate for specialization projection.}
\begin{wraptable}[7]{r}{0.38\textwidth}
\vspace{-0.3cm}
\renewcommand\arraystretch{0.85}
\setlength{\tabcolsep}{16pt}
\definecolor{backcolor}{RGB}{230, 239, 250}
\caption{Effects of varying learning rate for specialization projection.}
\label{Appendix::Tab::Ablation::lrratio}
\vspace{-1.5ex}
\begin{center}
\begin{normalsize}
\resizebox{1.0\linewidth}{!}{
\begin{tabular}{c|cc}
\toprule
learning rate   &  UCF      & HMDB      \\
    \specialrule{0.7pt}{0pt}{0pt}
$5 \times 10^{-6}$         & 83.1       & 52.8     \\  
\cellcolor{backcolor}$5 \times 10^{-5}$         & \cellcolor{backcolor}\textbf{83.7}       & \cellcolor{backcolor}\textbf{53.2}     \\    
$5 \times 10^{-4}$         & 82.6       & 52.3     \\   
\bottomrule
\end{tabular}
}
\end{normalsize}
\end{center}
\vskip -0.5cm
\end{wraptable}
We study the decoupling effect of different learning rates applied to projection parameters in Table~\ref{Appendix::Tab::Ablation::lrratio}.
Specifically, we fixed the learning rate applied to the CLIP parameters and adjusted the learning rate for the projection layer.
Results indicate that the learning rate ratio of around 10 offers the optimal results for space decoupling. Ratios below this threshold cannot effectively decouple the space, while higher ratios risk compromising the alignment between modalities.

\begingroup
\setlength{\parskip}{3pt plus 1pt minus 1pt}
\setlength{\abovedisplayskip}{4pt plus 1pt minus 1pt}
\setlength{\belowdisplayskip}{4pt plus 1pt minus 1pt}
\setlength{\abovedisplayshortskip}{2pt plus 1pt}
\setlength{\belowdisplayshortskip}{3pt plus 1pt minus 1pt}

\section{Theoretical Analysis of Relative Structure Distillation (RSD)}
\label{Appendix::Sec::Theory}

To better clarify the theoretical intuition behind RSD, we present a theoretical analysis to better support its underlying rationale.

\subsection{OOD Geometric Non-identifiability of Standard KL Distillation}

Standard KL distillation directly matches the teacher and student predictions on ID samples:
\[
\mathcal{L}_{\mathrm{KL}} = D_{\mathrm{KL}}\left(\sigma\left(v_t W_t^\top/\tau\right)\middle|\sigma\left(v_s W_s^\top/\tau\right)\right).
\]
Geometrically, each logit is the projection of a visual embedding onto an ID text direction. Let $\mathcal{U}=\operatorname{rowspan}(W_t)$ be the subspace spanned by the ID text embeddings and  $\operatorname{rank}(W_t)<D$. Standard KL therefore observes only the component of a visual embedding within $\mathcal{U}$:
\[
v_t W_t^\top=(v_t P_{\mathcal{U}})W_t^\top.
\]
where $P_{\mathcal{U}}$ denotes the orthogonal projector onto $\mathcal{U}$. Consequently, changes in the orthogonal complement $\mathcal{U}^\perp$ are invisible to the direct logits. To see this, keep the text embeddings fixed and consider a norm-preserving transformation of the visual space:
\[
v_s=v_t T,\qquad W_s=W_t,\qquad T=P_{\mathcal{U}}+A P_{\mathcal{U}^\perp},
\]
where $A$ is any orthogonal transformation on $\mathcal{U}^\perp$. Since $T$ leaves every ID text direction unchanged,
\[
v_s W_s^\top=v_t T W_t^\top=v_t W_t^\top.
\]
Thus, the visual geometry can change in $\mathcal{U}^\perp$ while all ID logits and the standard KL loss remain unchanged. However, an unseen text embedding may contain a component in $\mathcal{U}^\perp$, in which case
\[
v_s w_{\mathrm{OOD}}^\top\ne v_t w_{\mathrm{OOD}}^\top
\]
in general. Hence,
\[
\boxed{\text{identical ID logits}\not\Rightarrow\text{identical OOD cross-modal structure}.}
\]
The text branch admits a symmetric argument, and jointly updating both encoders introduces additional unconstrained degrees of freedom. This motivates geometric measurements beyond the directions covered by the ID texts.

\subsection{Random Geometric Anchors as Isotropic Probes}

Each anchor is sampled and normalized as
\[
g_m\sim\mathcal{N}(0,I),\qquad z_m=\frac{g_m}{||g_m||_2}.
\]
By rotational invariance, $z_m$ is uniformly distributed on the unit hypersphere and satisfies
\[
\mathbb{E}[z_m z_m^\top]=\frac{1}{D}I.
\]
For any representation change $e\in\mathbb{R}^{1\times D}$,
\[
\mathbb{E}_{z_m}\left[(e z_m)^2\right]=\frac{1}{D}||e||_2^2.
\]
Therefore, random anchors have equal expected sensitivity to every direction and introduce no systematic bias toward the ID subspace or any particular semantic direction.

For $M$ anchors, define
\[
S_Z:=Z^\top Z=\sum_{m=1}^{M}z_m z_m^\top,\qquad \mathbb{E}[S_Z]=\frac{M}{D}I.
\]
A finite $Z$ probes only finitely many directions in one iteration. However, $Z$ is isotropic in expectation, and repeated resampling continually introduces new probing directions throughout training. It thus provides a data-independent mechanism for probing the shared CLIP representation space.

\subsection{Additional Geometric Constraints Beyond Direct Logit Matching}

Let $w_{a,n}$ denote the $n$-th row of $W_a$, where $a\in\{t,s\}$, and define the symmetric visual-text relation matrix
\[
C_{a,n}:=\operatorname{sym}(v_a^\top w_{a,n})=\frac{1}{2}\left(v_a^\top w_{a,n}+w_{a,n}^\top v_a\right).
\]
A direct CLIP logit observes only its trace:
\[
v_a w_{a,n}^\top=\operatorname{tr}(C_{a,n}).
\]
Because softmax is invariant to a common logit shift, consider the teacher-student discrepancy for a class pair $(n,k)$:
\[
\Delta_{nk}:=(C_{s,n}-C_{s,k})-(C_{t,n}-C_{t,k}).
\]
Matching the direct logits constrains only
\[
\operatorname{tr}(\Delta_{nk})=0,
\]
which does not imply $\Delta_{nk}=0$. Thus, a realizable nonzero trace-free relation change can remain invisible to standard KL.

By contrast, the anchored logit is
\[
v_a Z^\top Z w_{a,n}^\top=\sum_{m=1}^{M}(v_a z_m)(z_m^\top w_{a,n}^\top)=\langle S_Z,C_{a,n}\rangle_F.
\]
RSD therefore measures the class-relation discrepancy as
\[
\langle S_Z,\Delta_{nk}\rangle_F=\sum_{m=1}^{M}z_m^\top\Delta_{nk}z_m.
\]
Writing
\[
S_Z=\frac{M}{D}I+E_Z,\qquad \mathbb{E}[E_Z]=0,
\]
gives
\[
\langle S_Z,\Delta_{nk}\rangle_F=\frac{M}{D}\operatorname{tr}(\Delta_{nk})+\langle E_Z,\Delta_{nk}\rangle_F.
\]
The first term is a scaled direct-logit measurement. The second term, induced by finite-sample anisotropic fluctuations, provides the additional geometric observation.

For $z\sim\mathrm{Unif}(\mathbb{S}^{D-1})$ and any symmetric $\Delta$,
\[
\mathbb{E}_z\left[(z^\top\Delta z)^2\right]=\frac{2||\Delta||_F^2+\operatorname{tr}(\Delta)^2}{D(D+2)}.
\]
In particular, if a nonzero relation change is invisible to direct logits, i.e., $\operatorname{tr}(\Delta_{nk})=0$, then
\[
\boxed{\mathbb{E}_Z\left[\langle Z^\top Z,\Delta_{nk}\rangle_F^2\right]=\frac{2M}{D(D+2)}||\Delta_{nk}||_F^2>0.}
\]
Hence, every nonzero discrepancy in the direct-logit null space produces a positive expected squared anchor response. Rather than providing semantics, the random anchors $Z$ act as directional probes that expose otherwise unobserved teacher–student geometric discrepancies.

\subsection{Implication for Suppressing OOD Alignment Shift}

The above analysis shows that standard KL leaves a geometric null space in which cross-modal relations can change without affecting ID logits, whereas resampled isotropic anchors make such changes observable with positive expected squared responses. 
Any anchor-measured logit-gap discrepancy changes the softmax distribution and is therefore penalized by RSD loss. 
Since OOD concepts share the pretrained CLIP space and may depend on directions unconstrained by ID logits, \textbf{RSD reduces the freedom for cross-modal distortion, thereby suppressing destructive drift of the pretrained cross-modal structure in OOD regions.}
\par
\endgroup

\section{More Analysis and Discussions}
\label{Appendix::Sec::Overlap}
\paragraph{Analysis of Random Geometric Anchors Overlapping with ID Space}
\begin{wrapfigure}{r}{0.38\textwidth} 
    \centering
    \vspace{-3pt} 
    \includegraphics[width=\linewidth]{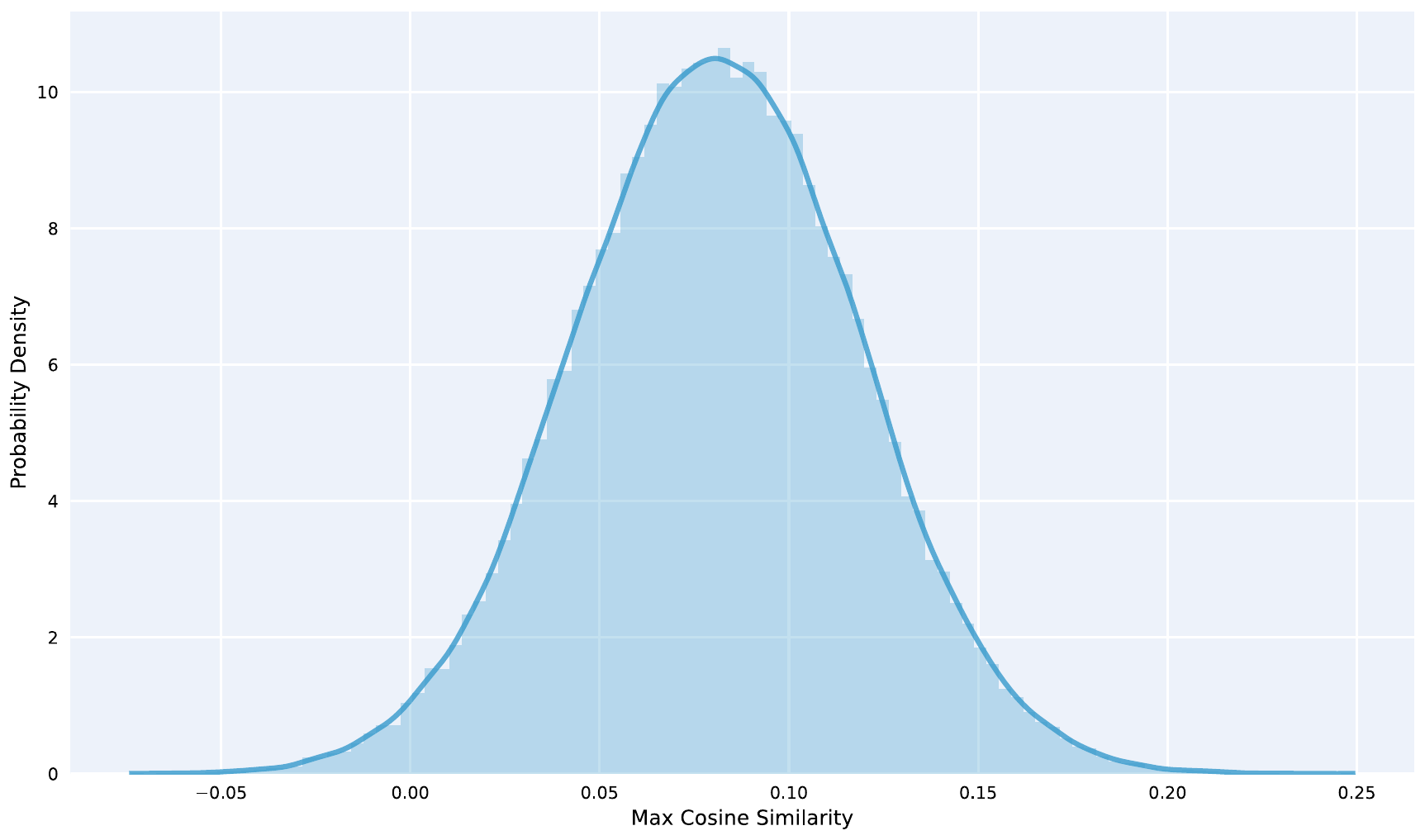}
    \vspace{-10pt} 
    \caption{Max cosine similarities between random geometric anchors and ID text representations}
    \label{Fig::overlap_dist}
    \vspace{-10pt} 
\end{wrapfigure}
As discussed in the main text, while our proposed random geometric anchors could theoretically overlap with the in-distribution training distribution, this probability is negligible during fine-tuning. This is due to the fact that CLIP’s representation space occupies a highly narrow conical region within the D-dimensional hypersphere. In Figure~\ref{Fig::overlap_dist}, we visualize the distribution of \textbf{maximum cosine similarities} between 10K randomly sampled geometric anchors and the CLIP training text representations. We observe that all similarity scores are below 0.25, indicating that no anchors overlap with the training distribution. This finding is consistent with the conclusions presented in \cite{MindTheGap}. Furthermore, we calculated the intrinsic dimension of the CLIP training text representations according to~\cite{IntrinsicDim}. The resulting dimension is 11, which is significantly lower than the embedding dimension of 512. This further confirms that the ID space occupies only a minuscule fraction of the 512-hypersphere.

\paragraph{Discussion on image-domain generalization.}
Although TACO is designed for open-vocabulary video adaptation, its benefit is not limited to the video domain. As shown in Table~\ref{Appendix::Tab::ImageDG}, TACO achieves the best performance on ImageNet (IN), ImageNet-V2, ImageNet-S, and ImageNet-R, while remaining competitive on IN-A. These results show that TACO improves not only in-domain accuracy but also robustness under distribution shift. This suggests that the gain of TACO does not mainly come from video-specific temporal modeling, but from preserving transferable cross-modal structure during adaptation.
By preserving the relative structure of the pretrained CLIP space and reducing overspecialization to the source data, TACO remains effective beyond video recognition, indicating its potential as a general adaptation framework for vision-language models.

\begin{table*}[t]
\renewcommand\arraystretch{0.9}
\setlength{\tabcolsep}{14pt}
\caption{Comparison with existing methods on image-domain generalization benchmarks.}
\vspace{-2.5ex}
\label{Appendix::Tab::ImageDG}
\definecolor{backcolor}{RGB}{230, 239, 250}
\begin{center}
\begin{normalsize}
\resizebox{0.8\linewidth}{!}{
\begin{tabular}{lccccc}
\toprule
Method & IN & IN-A & IN-V2 & IN-S & IN-R \\
\midrule
CoCoOp~\cite{cocoop}    & 71.02 & 50.63 & 64.07 & 48.75 & 76.18 \\
MaPLe~\cite{maple}     & 70.72 & 50.90 & 64.07 & 49.15 & 76.98 \\
PromptSRC~\cite{promptsrc} & 71.27 & 50.90 & 64.35 & 49.55 & \underline{77.80} \\
WISE-FT~\cite{wise-ft}   & \underline{72.38} & 51.07 & \underline{65.29} & \underline{49.72} & 72.48 \\
MMA~\cite{mma}       & 71.00 & \underline{51.12} & 64.33 & 49.13 & 77.32 \\
MMRL~\cite{mmrl}      & 72.03 & \textbf{51.20} & 64.47 & 49.17 & 77.13 \\
\cellcolor{backcolor}\textbf{TACO} &
\cellcolor{backcolor}\textbf{73.18} &
\cellcolor{backcolor}50.70 &
\cellcolor{backcolor}\textbf{65.55} &
\cellcolor{backcolor}\textbf{50.52} &
\cellcolor{backcolor}\textbf{78.20} \\
\bottomrule
\end{tabular}
}
\end{normalsize}
\end{center}
\vskip -0.3cm
\end{table*}

\paragraph{Discussion on the relation to LP-FT~\cite{LP-FT}.}
Compared with LP-FT, the key difference lies in the task setting and the resulting challenge. LP-FT studies single-modal transfer learning, where a pretrained vision encoder is adapted and evaluated under a fixed label space. In contrast, our work focuses on open-vocabulary adaptation in a CLIP-style joint visual-text embedding space, where the model is fine-tuned on known categories but evaluated on unseen ones. Under this setting, the main difficulty is not only preventing visual feature distortion, but also preserving cross-modal alignment in OOD space. Moreover, the motivation of LP-FT is rooted in the trade-off between linear probing and full fine-tuning for a single vision encoder, whereas our method is motivated by the inconsistency between fine-tuning and evaluation objectives in open-vocabulary adaptation. Based on this perspective, Section~\ref{Sec::Analysis} further provides observations that are specific to our setting, including the source of adapted OOD generalization and the relationship between alignment shift and OOD performance. Correspondingly, our technical solution is also fundamentally different: rather than mitigating the coupled update between a classification head and a visual encoder, TACO explicitly regularizes the relative structure of the joint embedding space and decouples the representation space from the optimization space to better preserve cross-modal consistency during adaptation.

\paragraph{Difference from FROSTER.}
While FROSTER~\cite{FROSTER} also employs a CLIP teacher to preserve generalization, our method differs primarily in design motivation and supervision scope.

FROSTER is motivated by preventing student features from overfitting to the training distribution and therefore separately constrains the absolute visual and textual features to remain close to their teacher counterparts, without explicitly modeling the cross-modal alignment between them. In contrast, RSD directly regularizes geometric drift in the OOD space, treating cross-modal alignment shift---rather than modality-wise feature deviation---as the key property to preserve for generalization. In terms of supervision scope, FROSTER matches teacher and student features only on training samples, confining its supervision to the ID distribution. RSD instead explicitly extends supervision into the OOD space---a region FROSTER cannot reach by construction.

\paragraph{Difference from relation-based distillation.}
While conceptually similar to relation-based distillation methods (e.g., CRCD~\cite{CRCD}), our method differs primarily in what relations are constructed and where they are anchored.

These methods aim to enhance discriminative organization among training categories by preserving the relative structure among training samples, matching sample-to-sample relations (e.g., via NCE-style objectives) computed entirely within the training batch---so their relational supervision remains confined to the ID distribution. In contrast, RSD constructs relations between ID embeddings and anchors explicitly placed beyond the ID manifold, allowing it to capture ID-to-OOD relational shifts, regardless of how the relations among training samples are defined. This distinction is empirically supported by Table~\ref{Tab::AS::OOD}, supporting that the gain comes specifically from anchoring relations in a constructed OOD space, rather than from relation matching itself.

\paragraph{Metrics for structural preservation on unseen categories.}
For the comparison in Table~\ref{Tab::UnseenStructure}, all three metrics use the original CLIP space as a reference.
\textbf{Text-direction cosine} is the mean cosine similarity between corresponding pairwise unseen-class directions $w_i-w_j$ in the adapted model and CLIP, where $w_i$ denotes the text embedding of class $i$.
\textbf{Text kNN overlap@5} is the average proportion of each unseen class's five nearest text neighbors shared with CLIP.
\textbf{Cross-modal Spearman correlation} is the mean per-video rank correlation between the adapted model's similarities to all unseen-class texts and the corresponding CLIP similarities.
Higher values indicate better preservation of class directions, local text neighborhoods, and cross-modal rankings, respectively.

\begin{figure*}[t]
\begin{center}
\centerline{
    \includegraphics[width=0.75\linewidth]{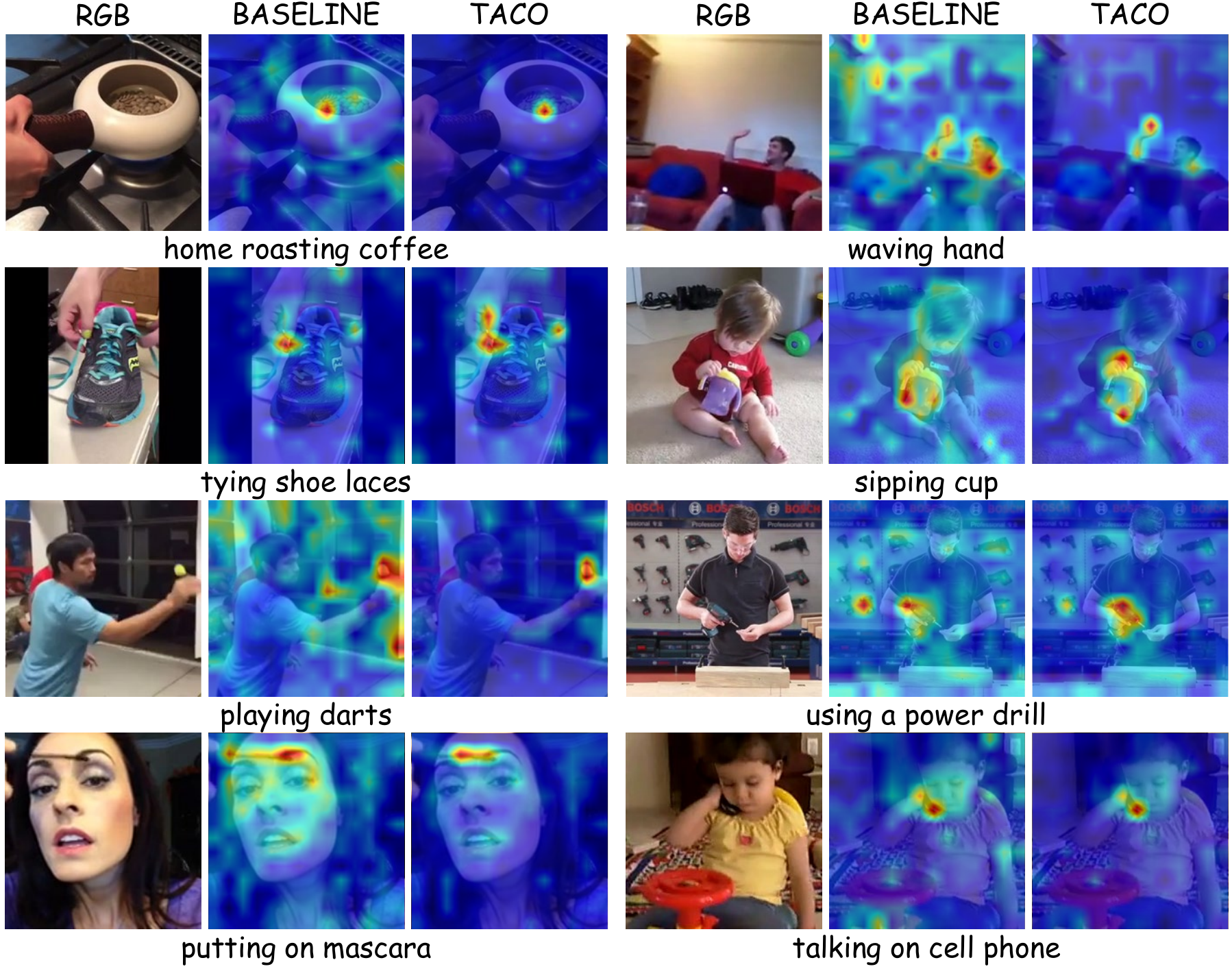}}
\vspace{-4pt}
\caption{Visualization of model attention maps.
}
\label{Fig::heatmap}
\end{center}
\vskip -1.0cm
\end{figure*}

%
%
%
%
\section{Visualizations}
\label{Appendix::Sec::Visualization}

\paragraph{Model Attention.}
To better understand what generalization knowledge the model captures when processing videos of unknown categories, we visualize the attention heatmap of the model in Figure~\ref{Fig::heatmap}.
The attention maps are calculated between the text embedding and the patch embeddings output by the model.
As shown in the figure, our model can focus more intently on key objects or actions, with less distraction from the background or other irrelevant objects.
This phenomenon suggests that our model can learn more generalized video knowledge while maintaining the strong spatial understanding capabilities of the original CLIP during fine-tuning.

%
%
\begin{table}[ht]
\renewcommand\arraystretch{1.0}
\caption{Textual prompt templates of TACO.}
\definecolor{midgrey}{RGB}{225,225,225}
\vskip 0.15in
\centering
\scalebox{1.0}{
\begin{tabular}{l} \toprule
    \cellcolor{midgrey}Templates \\ \midrule
    'a photo of \{category\}.'\\
    'a photo of a person \{category\}.'\\
    'a photo of a person using \{category\}.'\\
    'a photo of a person doing \{category\}.'\\
    'a photo of a person during \{category\}.'\\
    'a photo of a person performing \{category\}.'\\
    'a photo of a person practicing \{category\}.'\\
    'a video of \{category\}.'\\
    'a video of a person \{category\}.'\\
    'a video of a person using \{category\}.'\\
    'a video of a person doing \{category\}.'\\
    'a video of a person during \{category\}.'\\
    'a video of a person performing \{category\}.'\\
    'a video of a person practicing \{category\}.'\\
    'a example of \{category\}.'\\
    'a example of a person \{category\}.'\\
    'a example of a person using \{category\}.'\\
    'a example of a person doing \{category\}.'\\
    'a example of a person during \{category\}.'\\
    'a example of a person performing \{category\}.'\\
    'a example of a person practicing \{category\}.'\\
    'a demonstration of \{category\}.'\\
    'a demonstration of a person \{category\}.'\\
    'a demonstration of a person using \{category\}.'\\
    'a demonstration of a person doing \{category\}.'\\
    'a demonstration of a person during \{category\}.'\\
    'a demonstration of a person performing \{category\}.'\\
    'a demonstration of a person practicing \{category\}.'\\
    \bottomrule
\end{tabular}}
\label{Appendix::Tab::templates}
\end{table}

\section{Textual Prompts Used in Evaluation}
\label{Appendix::Sec::TPT}
Following the previous work~\citep{MoTE}, we adopt a set of hand-craft textual prompt templates to generate text embeddings during the evaluations. 
Following CLIP~\citep{CLIP}, we perform prompt ensembling over the 28 templates in order to provide comprehensive semantics, as listed in Table \ref{Appendix::Tab::templates}.

\section{Dataset Details}
\label{Appendix::Sec::Dataset}
\paragraph{Kinetics-400~\citep{K400}} is a large-scale dataset in the video domain. 
The dataset contains $\sim$240k training videos and $\sim$20k validation videos in 400 human action categories, with an average length of 10 seconds. 
The high quality of the dataset makes it the most popular benchmark for video recognition

\paragraph{Kinetics-600~\citep{K600}} is an extension of Kinetics-400, consisting of $\sim$392k training videos, $\sim$30k validation videos, and $\sim$60k test videos in 600 human action categories.
The dataset contains an additional 220 new action categories over Kinetics-400.
We evaluate the zero-shot performance on 220 new categories and adopt three splits provided by the previous work~\citep{ER}.
We use its test set for evaluation and report the average performance on three splits.

\paragraph{UCF-101~\citep{UCF101}} is an action recognition dataset that contains 13,320 videos in 101 action categories, collected from YouTube. There are three official splits of training data and validation data.

\paragraph{HMDB-51~\citep{HMDB51}} contains 7,000 videos in 51 action categories, collected from movie clips and web videos. There are three official splits of the dataset, each with 3,570 training data and 1,530 validation data samples.

\paragraph{Something-Something V2~\citep{SSv2}} is a temporal-heavy dataset that requires the fine-grained temporal understanding capability of the model. It contains 220,000 videos in 174 action categories. 

\newpage







\end{document}